\documentclass{article}
\usepackage{selectp}

\usepackage{microtype}
\usepackage{graphicx}
\usepackage{subfigure}
\usepackage{booktabs} %
\usepackage[table]{xcolor}
\usepackage{mfirstuc}
\usepackage{xspace}
\usepackage{multirow}
\usepackage{glossaries}
\usepackage[normalem]{ulem}
\usepackage{titlesec}
\usepackage{floatrow}
\usepackage{dsfont}
\newfloatcommand{capbtabbox}{table}[][\FBwidth]

\usepackage{hyperref}

\usepackage[arxiv]{icml2021}

\usepackage{amsmath,amsfonts,bm}

\def\eqref#1{equation~\ref{#1}}

\def\1{\bm{1}}

\def\vtheta{{\bm{\theta}}}
\def\vdelta{{\bm{\delta}}}
\def\va{{\bm{a}}}

\def\vx{{\bm{x}}}

\DeclareMathAlphabet{\mathsfit}{\encodingdefault}{\sfdefault}{m}{sl}
\SetMathAlphabet{\mathsfit}{bold}{\encodingdefault}{\sfdefault}{bx}{n}

\def\gX{{\mathcal{X}}}
\def\gY{{\mathcal{Y}}}

\def\sA{{\mathbb{A}}}

\def\sD{{\mathbb{D}}}

\def\sS{{\mathbb{S}}}

\newcommand{\E}{\mathbb{E}}

\DeclareMathOperator*{\argmin}{arg\,min}

\DeclareMathOperator{\sign}{sign}

\DeclareMathOperator*{\maximize}{max}

\DeclareMathOperator{\proj}{proj}

\newacronym{pgd}{PGD}{Projected Gradient Descent}
\newacronym{bim}{BIM}{Basic Iterative Method}
\newacronym{fgsm}{FGSM}{Fast Gradient Sign Method}
\newacronym{wrn}{\textsc{Wrn}}{Wide ResNet}
\newacronym{sgd}{SGD}{Stochastic Gradient Descent}
\newacronym{ddpm}{DDPM}{Denoising Diffusion Probabilistic Model}
\newacronym{vdvae}{VDVAE}{Very Deep Variational Auto-Encoder}
\newacronym{fid}{FID}{Frechet Inception Distance}
\newacronym{is}{IS}{Inception Score}
\newacronym{gan}{GAN}{Generative Adversarial Network}
\newacronym{vae}{VAE}{Variational AutoEncoder}

\newcommand{\cifar}{\textsc{Cifar-10}\xspace}

\newcommand{\cifarh}{\textsc{Cifar-100}\xspace}
\newcommand{\tinyimages}{\textsc{80M-Ti}\xspace}
\newcommand{\imagenet}{\textsc{ImageNet}\xspace}

\newcommand{\svhn}{\textsc{Svhn}\xspace}
\newcommand{\linf}{\ensuremath{\ell_\infty}\xspace}
\newcommand{\ltwo}{\ensuremath{\ell_2}\xspace}
\newcommand{\lp}{\ensuremath{\ell_p}\xspace}
\newcommand{\autoattack}{\textsc{AutoAttack}\xspace}
\newcommand{\autopgd}{\textsc{AutoPgd}\xspace}
\newcommand{\multitargeted}{\textsc{MultiTargeted}\xspace}

\newcommand{\pgd}[1]{\textsc{Pgd}\textsuperscript{$#1$}\xspace}
\newcommand{\xent}{l_{\textrm{ce}}}
\newcommand{\wrn}{\gls*{wrn}\xspace}

\definecolor{TartOrange}{HTML}{ff2e35}
\definecolor{Orange}{HTML}{ff7825}
\definecolor{Mango}{HTML}{ffc013}
\definecolor{AppleGreen}{HTML}{7cb81b}
\definecolor{Blue}{HTML}{1173b0}
\definecolor{BdazzledBlue}{HTML}{2e58a5}
\definecolor{Purple}{HTML}{5b3590}
\definecolor{Sunglow}{HTML}{FFCA3A}

\definecolor{header}{gray}{0.9}
\definecolor{subheader}{rgb}{0.63, 0.79, 0.95}
\newcommand{\Tstrut}{\rule{0pt}{2.6ex}}
\newcommand{\Bstrut}{\rule[-0.9ex]{0pt}{0pt}}
\newcommand{\TBstrut}{\Tstrut\Bstrut}

\newcommand{\squishlist}{
   \begin{list}{$\bullet$}
    { \setlength{\itemsep}{0pt}      \setlength{\parsep}{3pt}
      \setlength{\topsep}{3pt}       \setlength{\partopsep}{0pt}
      \setlength{\leftmargin}{1.5em} \setlength{\labelwidth}{1em}
      \setlength{\labelsep}{0.5em} } }

\newcommand{\squishend}{
    \end{list}  }

\icmltitlerunning{Fixing Data Augmentation to Improve Adversarial Robustness}

\begin{document}

\twocolumn[

\icmltitle{Fixing Data Augmentation to Improve Adversarial Robustness}

\icmlsetsymbol{equal}{*}

\begin{icmlauthorlist}
\icmlauthor{Sylvestre-Alvise Rebuffi}{equal,dm}
\icmlauthor{Sven Gowal}{equal,dm}
\icmlauthor{Dan A. Calian}{dm}
\icmlauthor{Florian Stimberg}{dm}
\icmlauthor{Olivia Wiles}{dm}
\icmlauthor{Timothy Mann}{dm}
\end{icmlauthorlist}

\icmlaffiliation{dm}{DeepMind, London, UK}

\icmlcorrespondingauthor{Sylvestre-Alvise Rebuffi}{sylvestre@google.com}
\icmlcorrespondingauthor{Sven Gowal}{sgowal@google.com}

\icmlkeywords{Robustness, Adversarial, Machine Learning, ICML}

\vskip 0.3in
]

\printAffiliationsAndNotice{\icmlEqualContribution} %

\begin{abstract}
Adversarial training suffers from \emph{robust overfitting}, a phenomenon where the robust test accuracy starts to decrease during training.
In this paper, we focus on both heuristics-driven and data-driven augmentations as a means to reduce robust overfitting.
First, we demonstrate that, contrary to previous findings, when combined with model weight averaging, data augmentation can significantly boost robust accuracy.
Second, we explore how state-of-the-art generative models can be leveraged to artificially increase the size of the training set and further improve adversarial robustness.
Finally, we evaluate our approach on \cifar against \linf and \ltwo norm-bounded perturbations of size $\epsilon = 8/255$ and $\epsilon = 128/255$, respectively.
We show large absolute improvements of +7.06\% and +5.88\% in robust accuracy compared to previous state-of-the-art methods.
In particular, against \linf norm-bounded perturbations of size $\epsilon = 8/255$, our model reaches 64.20\% robust accuracy without using any external data, beating most prior works that use external data.
\textbf{Since its original publication (2 Mar 2021), this paper has been accepted to NeurIPS 2021 as two separate and updated papers~\citep{rebuffi2021data,gowal2021improving}. The new papers improve results and clarity.}
\end{abstract}

\section{Introduction}

Despite their success, neural networks are not intrinsically robust.
In particular, it has been shown that the addition of imperceptible deviations to the input, called adversarial perturbations, can cause neural networks to make incorrect predictions with high confidence \citep{carlini_adversarial_2017,carlini_towards_2017,goodfellow_explaining_2014,kurakin_adversarial_2016,szegedy_intriguing_2013}.
Starting with \citet{szegedy_intriguing_2013}, there has been a lot of work on understanding and generating adversarial perturbations \citep{carlini_towards_2017,athalye_synthesizing_2017}, and on building defenses that are robust to such perturbations \citep{goodfellow_explaining_2014,papernot_distillation_2015,madry_towards_2017,kannan_adversarial_2018}.
Among successful defenses are robust optimization techniques like the one developed by \citet{madry_towards_2017} that learn robust models by finding worst-case adversarial perturbations at each training step. %
In fact, adversarial training as proposed by \citeauthor{madry_towards_2017} is so effective~\citep{gowal_uncovering_2020} that it is the de facto standard for training adversarially robust neural networks.

\begin{figure}[t]
\begin{center}
\centerline{\includegraphics[width=.85\columnwidth]{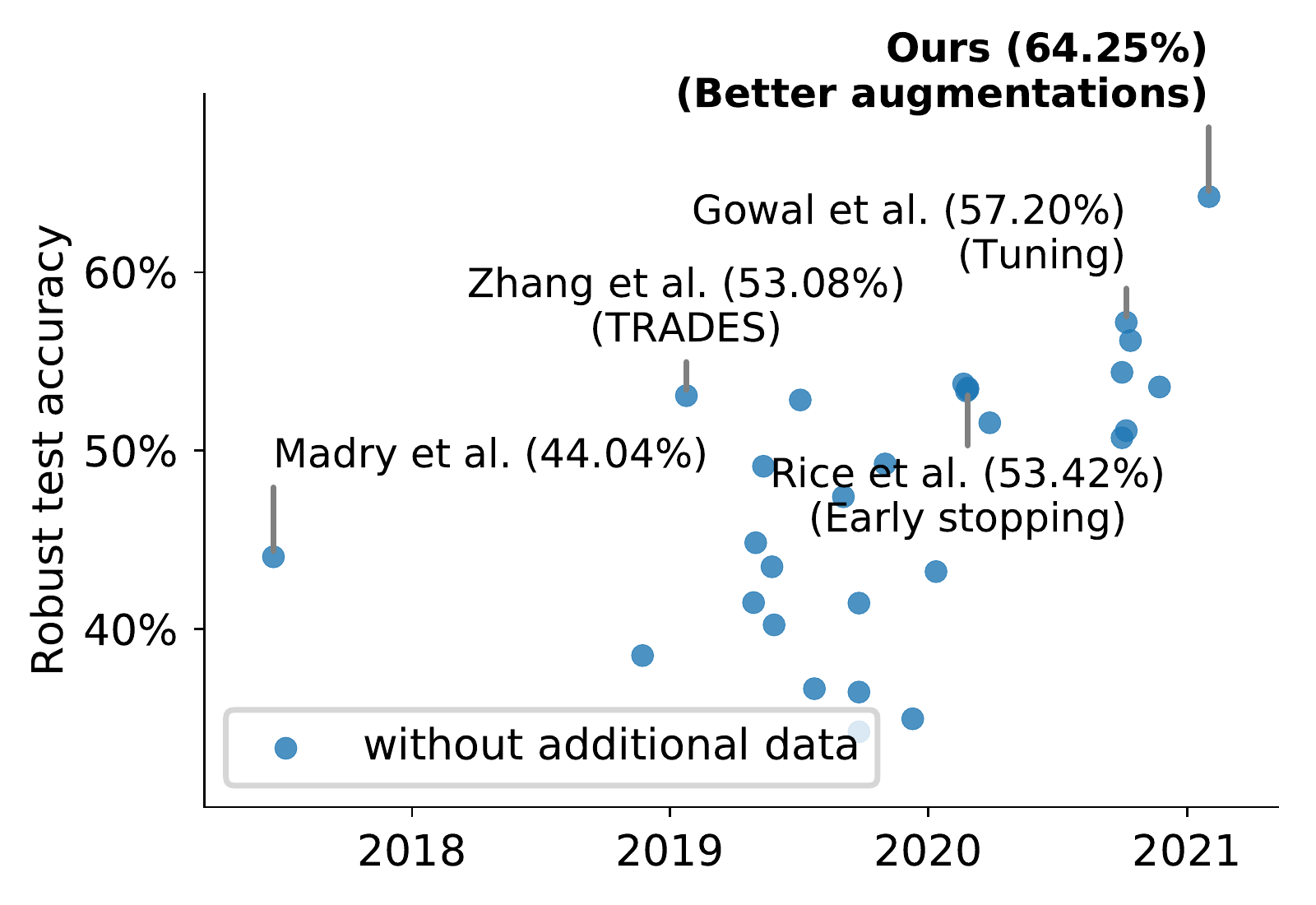}}
\caption{Robust accuracy of models against \autoattack~\citep{croce_reliable_2020} on \cifar with \linf perturbations of size $8/255$ displayed in publication order.
Our method builds on \citet{gowal_uncovering_2020} (shown above with 57.20\%) and explores how augmented and generated data can be used to improve robust accuracy by +7.05\% without using any additional external data (against \textsc{AA+MT}, the improvement is +7.06\%).
This constitutes the largest jump in robust accuracy in this setting in the past 2 years.}
\label{fig:history}
\end{center}
\vspace{-1cm}
\end{figure}

Since \citet{madry_towards_2017}, various modification to their original implementation have been proposed \citep{zhang_theoretically_2019,xie_feature_2018,pang_boosting_2020,huang_self-adaptive_2020,qin_adversarial_2019,rice_overfitting_2020, gowal_uncovering_2020}.
However, as shown in \autoref{fig:history}, improvements in robust accuracy brought by these newer techniques has dramatically slowed down in the past two years.
In a bid to improve robustness, \citet{hendrycks_using_2019,carmon_unlabeled_2019,uesato_are_2019,zhai_adversarially_2019,najafi_robustness_2019} pioneered the use of additional data (in both fully-supervised and semi-supervised settings).
\citet{gowal_uncovering_2020} (who trained the best currently available models on \cifar against \linf perturbations of size $\epsilon = 8/255$) could obtain a robust accuracy of 65.87\% when using additional unlabeled data, compared to 57.14\% without this data.
The gap between these settings motivates this paper.
In particular, we ask ourselves, \emph{``is is possible to make a better utilization of the original training set?''}
By making the observation that model weight averaging (WA)~\citep{izmailov_averaging_2018} helps robust generalization to a wider extent when robust overfitting is minimized, we propose to combine the data sampled from generative models (e.g., \citealp{ho2020denoising}) with data augmentation techniques such as \emph{Cutout}~\cite{devries2017improved}.
Overall, we make the following contributions:

\squishlist
\item We demonstrate how, when combined with model weight averaging, heuristics-driven augmentation techniques such as \emph{Cutout}, \emph{CutMix}~\cite{yun2019cutmix} and \emph{MixUp}~\cite{zhang2017mixup} can improve robustness.
\item To the contrary of \citet{rice_overfitting_2020,wu2020adversarial,gowal_uncovering_2020} which all tried data augmentation techniques without success, we are able to use any of these three aforementioned techniques to obtain new state-of-the-art robust accuracies. We find \emph{CutMix} to be the most effective method by reaching a robust accuracy of 60.07\% on \cifar against \linf perturbations of size $\epsilon = 8/255$ (an improvement of +2.93\% upon the state-of-the-art).
\item We explain how data-driven augmentations help improve diversity and complement heuristics-driven augmentations. We leverage additional generated inputs (i.e., inputs generated by generative models solely trained on the original data), and study three recent generative models: the \gls{ddpm} \citep{ho2020denoising}, the \gls{vdvae} \citep{child2021vdvae} and BigGAN~\citep{brock2018large}.
\item We show that images generated by \gls{ddpm} are more informative and allow us to reach a robust accuracy of 63.58\% on \cifar against \linf perturbations of size $\epsilon = 8/255$ (an improvement of +6.44\% upon SOTA).
\item Finally, we combine both heuristics- and data-driven approaches. On \cifar, we train models with 64.20\% and 80.38\% robust accuracy against \linf and \ltwo norm-bounded perturbations of size $\epsilon = 8/255$ and $\epsilon = 128/255$, respectively. Those are improvements of +7.06\% and +5.88\% upon state-of-the-art methods that do not make use of additional data. Notably, our best \cifar models beat all techniques that use additional data, except for the work by \citet{gowal_uncovering_2020}.
\squishend

\section{Related Work}

\begin{figure*}[t]
\centering
\subfigure[Adversarial training with and without additional external data from \tinyimages]{\label{fig:no_wa_external_vs_original}\includegraphics[width=0.3\textwidth]{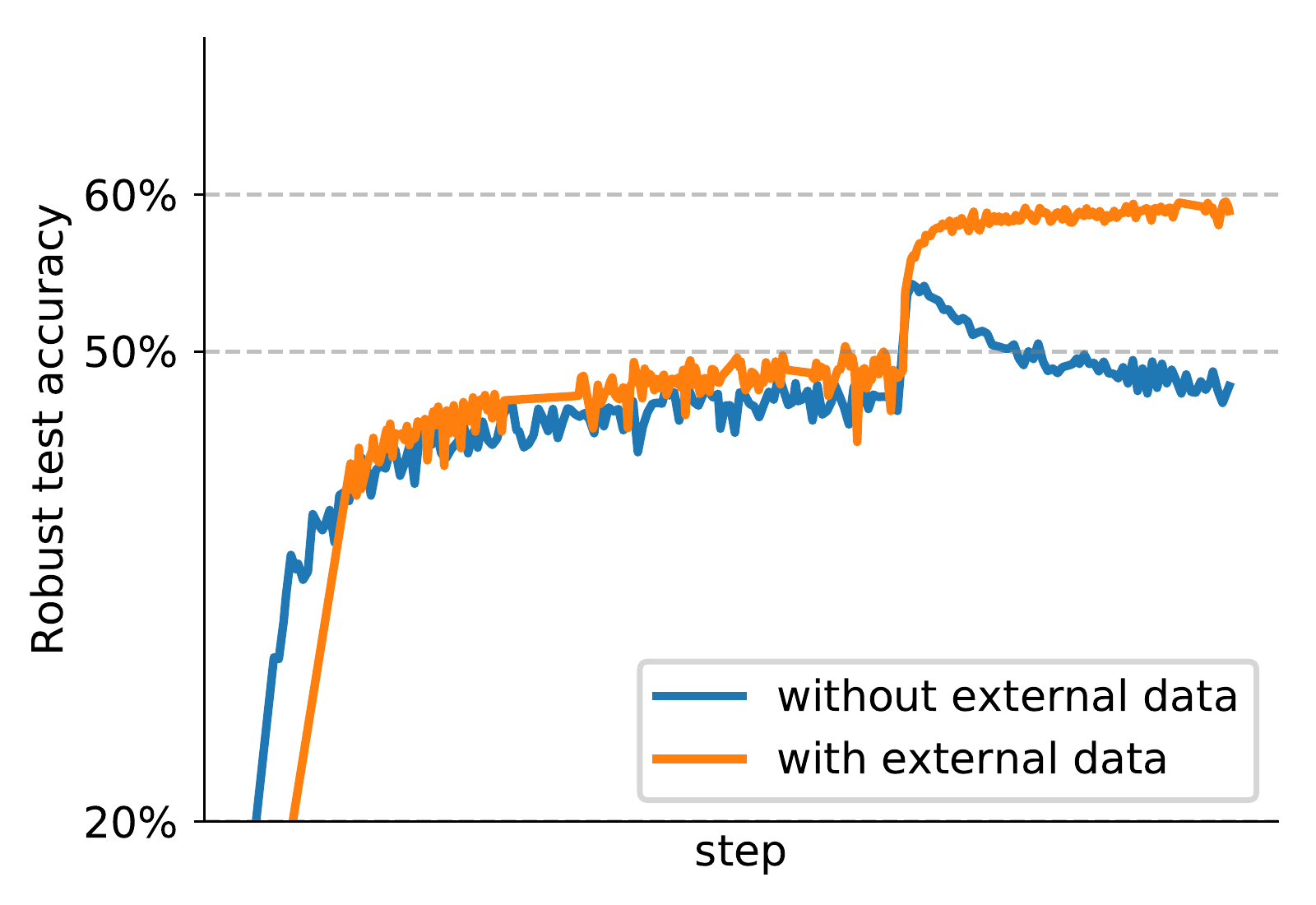}}
\hspace{.5cm}
\subfigure[Effect of WA without external data]{\label{fig:wa_vs_no_wa_robust_overfitting}\includegraphics[width=0.3\textwidth]{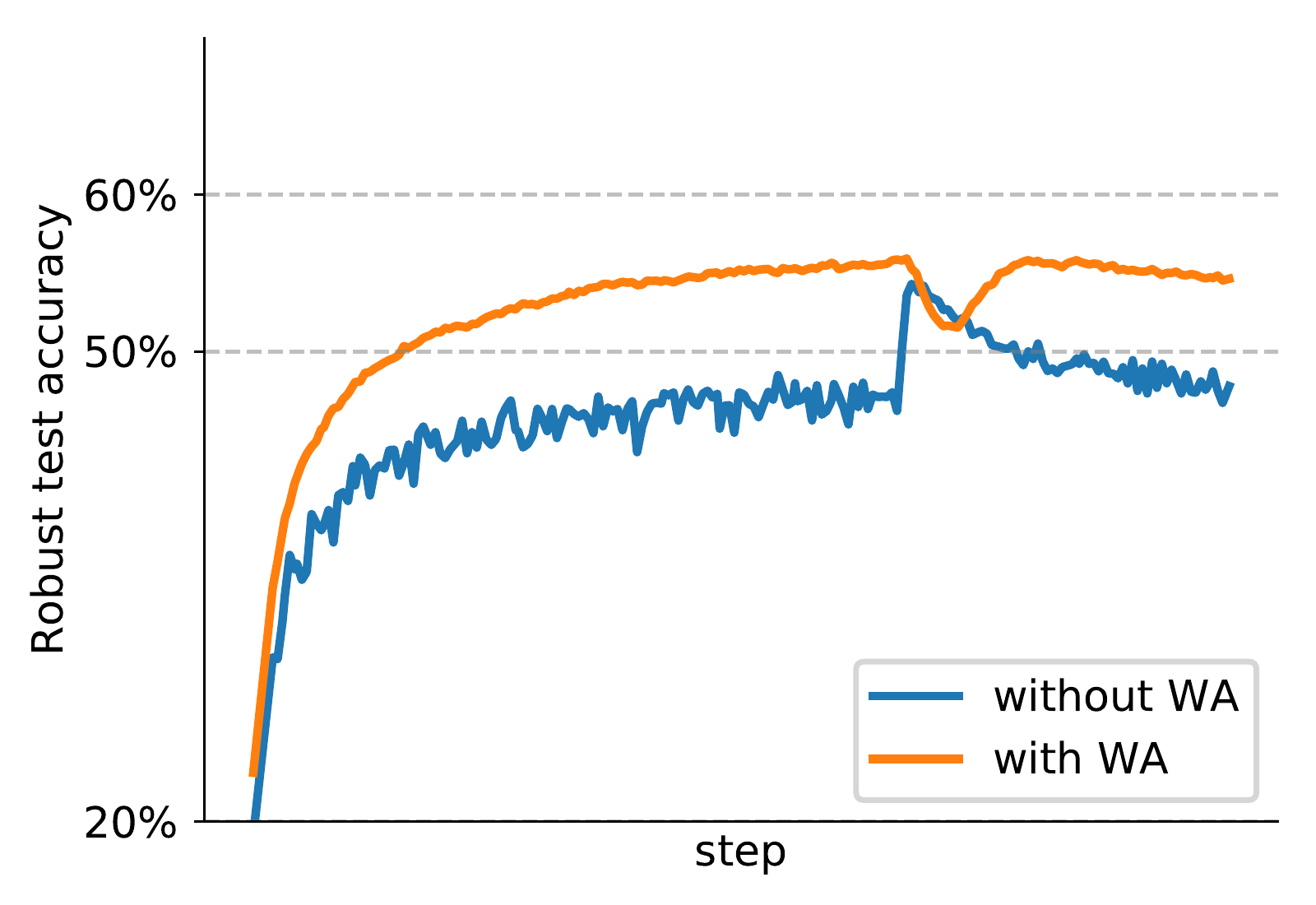}}
\hspace{.5cm}
\subfigure[Effect of WA with external data]{\label{fig:wa_vs_no_wa_external_data}\includegraphics[width=0.3\textwidth]{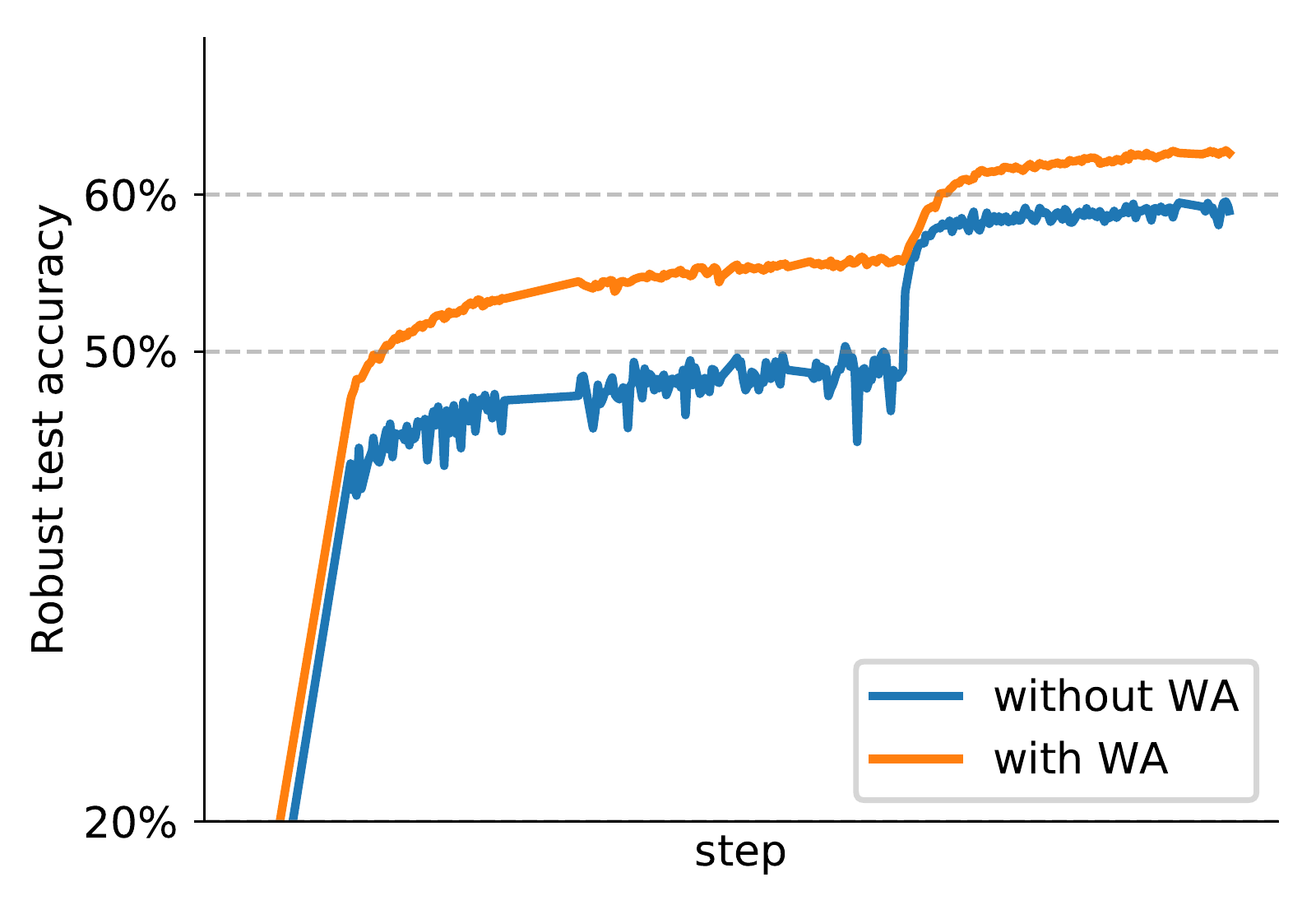}}
\caption{We compare the robust accuracy against $\epsilon_\infty = 8/255$ on \cifar of an adversarially trained \wrn-28-10.
Panel~\subref{fig:no_wa_external_vs_original} shows the impact of using additional external data from \tinyimages~\citep{80m} and illustrates \emph{robust overfitting}.
Panel~\subref{fig:wa_vs_no_wa_robust_overfitting} shows the benefit of \emph{model weight averaging} (WA) despite robust overfitting.
Panel~\subref{fig:wa_vs_no_wa_external_data} shows that WA remains effective and useful even when robust overfitting disappears.
The model is a \wrn-28-10 and all the panels show the evolution of the robust accuracy as training progresses (against \pgd{40}).
The jump in robust accuracy two-thirds through training is due to a drop in learning rate.
\label{fig:robust_overfitting}
}
\vspace{-.3cm}
\end{figure*}

\paragraph{Adversarial \lp-norm attacks.}

Since \citet{szegedy_intriguing_2013} observed that neural networks which achieve high accuracy on test data are highly vulnerable to adversarial examples, the art of crafting increasingly sophisticated adversarial examples has received a lot of attention.
\citet{goodfellow_explaining_2014} proposed the \gls{fgsm} which generates adversarial examples with a single normalized gradient step.
It was followed by R+\gls{fgsm} \citep{tramer_ensemble_2017}, which adds a randomization step, and the \gls{bim} \citep{kurakin_adversarial_2016}, which takes multiple smaller gradient steps.

\paragraph{Adversarial training as a defense.}

The adversarial training procedure~\citep{madry_towards_2017} feeds adversarially perturbed examples back into the training data.
It is widely regarded as one of the most successful method to train robust deep neural networks.
It has been augmented in different ways -- with changes in the attack procedure (e.g., by incorporating momentum; \citealp{dong_boosting_2017}), loss function (e.g., logit pairing; \citealp{mosbach_logit_2018}) or model architecture (e.g., feature denoising; \citealp{xie_feature_2018}).
Another notable work by \citet{zhang_theoretically_2019} proposed TRADES, which balances the trade-off between standard and robust accuracy, and achieved state-of-the-art performance against \linf norm-bounded perturbations on \cifar.
More recently, the work from \citet{rice_overfitting_2020} studied \emph{robust overfitting} and demonstrated that improvements similar to TRADES could be obtained more easily using classical adversarial training with early stopping.
This later study revealed that early stopping was competitive with many other regularization techniques and demonstrated that data augmentation schemes beyond the typical \emph{random padding-and-cropping} were ineffective on \cifar.
Finally, \citet{gowal_uncovering_2020} highlighted how different hyper-parameters (such as network size and model weight averaging) affect robustness.
They were able to obtain models that significantly improved upon the state-of-the-art, but lacked a thorough investigation on data augmentation schemes.
Similarly to \citet{rice_overfitting_2020}, they also make the conclusion that data augmentations beyond \emph{random padding-and-cropping} do not improve robustness.

\paragraph{Heuristics-driven data augmentation.}

Data augmentation has been shown to reduce the generalization error of standard (non-robust) training.
For image classification tasks, random flips, rotations and crops are commonly used~\cite{he2015deep}.
More sophisticated techniques such as \emph{Cutout}~\cite{devries2017improved} (which produces random occlusions), \emph{CutMix}~\cite{yun2019cutmix} (which replaces parts of an image with another) and \emph{MixUp}~\cite{zhang2017mixup} (which linearly interpolates between two images) all demonstrate extremely compelling results.
As such, it is rather surprising that they remain ineffective when training adversarially robust networks.

\paragraph{Data-driven data augmentation.}

Work, such as \emph{AutoAugment} \citep{cubuk_autoaugment:_2018} and related \emph{RandAugment} \citep{cubuk2019randaugment}, learn augmentation policies directly from data.
These methods are tuned to improve standard classification accuracy and have been shown to work well on \cifar, \cifarh, \svhn and \imagenet.
DeepAugment \citep{hendrycks_many_2020} explores how perturbations of the parameters of several image-to-image models can be used to generate augmented datasets that provide increased robustness to common corruptions \citep{hendrycks2018benchmarking}.
Similarly, generative models can be used to create novel views of images~\citep{plumerault2019controlling,jahanian2019steerability,harkonen_ganspace_2020} by manipulating them in latent space.
When optimized and used during training, these novel views reduce the impact of spurious correlations and improve accuracy~\citep{gowal_achieving_2019,wong2020learning}.
However, to the best of our knowledge, there is little \citep{madaan2020learning} to no evidence that generative models can be used to improve adversarial robustness against \lp-norm attacks.
In fact, generative models mostly lack diversity and it is widely believed that the samples they produce cannot be used to train classifiers to the same accuracy than those trained on original datasets \citep{ravuri_classification_2019}.

\section{Preliminaries and Hypothesis}
\label{sec:hypothesis}

\paragraph{Adversarial training.}

\citet{madry_towards_2017} formulate a saddle point problem to find model parameters $\vtheta$ that minimize the adversarial risk:
\begin{equation}
\argmin_\vtheta \E_{(\vx,y) \sim \mathcal{D}} \left[ \maximize_{\vdelta \in \sS} l(f(\vx + \vdelta; \vtheta), y) \right]
\label{eq:adversarial_risk}
\end{equation}
\noindent where $\mathcal{D}$ is a data distribution over pairs of examples $\vx$ and corresponding labels $y$, $f(\cdot; \vtheta)$ is a model parametrized by $\vtheta$, $l$ is a suitable loss function (such as the $0-1$ loss in the context of classification tasks), and $\sS$ defines the set of allowed perturbations.
For $\ell_p$ norm-bounded perturbations of size $\epsilon$, the adversarial set is defined as $\sS_p = \{ \vdelta ~|~ \| \vdelta \|_p \leq \epsilon \}$.
In the rest of this manuscript, we will use $\epsilon_p$ to denote $\ell_p$ norm-bounded perturbations of size $\epsilon$ (e.g., $\epsilon_\infty = 8/255$).
To solve the inner optimization problem, \citet{madry_towards_2017} use \gls*{pgd}, which replaces the non-differentiable $0-1$ loss $l$ with the cross-entropy loss $\xent$ and computes an adversarial perturbation $\hat{\vdelta} = \vdelta^{(K)}$ in $K$ gradient ascent steps of size $\alpha$ as
\begin{equation}
\resizebox{\linewidth}{!}{
$\vdelta^{(k+1)} \gets \proj_{\sS} \left( \vdelta^{(k)} + \alpha \sign \left(\nabla_{\vdelta^{(k)}} \xent(f(\vx + \vdelta^{(k)}; \vtheta), y) \right)\right)$
\label{eq:bim}
}
\end{equation}
where $\vdelta^{(0)}$ is chosen at random within $\sS$, and where $\proj_{\sA}(\va)$ projects a point $\va$ back onto a set $\sA$, $\proj_{\sA}(\va) = \mathrm{argmin}_{\va' \in \sA} \|\va - \va'\|_2$.
We will refer to this inner optimization procedure with $K$ steps as \pgd{K}.

\paragraph{Robust overfitting.}

To the contrary of standard training, which often shows no \emph{overfitting} in practice~\citep{zhang2017understanding}, adversarial training suffers from \emph{robust overfitting}~\citep{rice_overfitting_2020}.
Robust overfitting is the phenomenon by which robust accuracy on the test set quickly degrades while it continues to rise on the train set (clean accuracy on both sets continues to improve as well).
\citet{rice_overfitting_2020} propose to use early stopping as the main contingency against robust overfitting, and demonstrate that it also allows to train models that are more robust than those trained with other regularization techniques (such as data augmentation or increased \ltwo-regularization).
They observed that some of these other regularization techniques could reduce the impact of overfitting at the cost of producing models that are over-regularized and lack overall robustness and accuracy.
There is one notable exception which is the addition of external data \citep{carmon_unlabeled_2019,uesato_are_2019}.
\autoref{fig:no_wa_external_vs_original} shows how the robust accuracy (evaluated on the test set) evolves as training progresses on \cifar against $\epsilon_\infty = 8/255$.
Without external data, robust overfitting is clearly visible and appears shortly after the learning rate is dropped (the learning rate is decayed by 10$\times$ two-thirds through training in  a schedule is similar to \citealp{rice_overfitting_2020} and commonly used since \citealp{madry_towards_2017}).
Robust overfitting completely disappears when an additional set of 500K pseudo-labeled images from \tinyimages~\citep{80m} is introduced.

\paragraph{Model weight averaging.}

Model weight averaging (WA)~\citep{izmailov_averaging_2018} can be implemented using an exponential moving average $\vtheta'$ of the model parameters $\vtheta$ with a decay rate $\tau$ (i.e., $\vtheta' \gets \tau \cdot \vtheta' + (1 - \tau) \cdot \vtheta$ at each training step).
During evaluation, the weighted parameters $\vtheta'$ are used instead of the trained parameters $\vtheta$.
\citet{gowal_uncovering_2020,chen2021robust} discovered that model weight averaging can significantly improve robustness on a wide range of models and datasets.
\citet{chen2021robust} argue (similarly to \citealp{wu2020adversarial}) that WA leads to a flatter adversarial loss landscape, and thus a smaller robust generalization gap.
\citet{gowal_uncovering_2020} also explain that, in addition to improved robustness, WA reduces sensitivity to early stopping.
While this is true, it is important to note that WA is still prone to robust overfitting.
This is not surprising, since the exponential moving average ``forgets'' older model parameters as training goes on.
\autoref{fig:wa_vs_no_wa_robust_overfitting} shows how the robust accuracy evolves as training progresses when using WA.
We observe that, after the change of learning rate, the averaged weights are increasingly affected by overfitting, thus resulting in worse robust accuracy for the averaged model.

\paragraph{Hypothesis.}

As WA results in flatter, wider solutions compared to the steep decrease in robust accuracy observed for \gls{sgd}~\citep{chen2021robust}, it is natural to ask ourselves whether WA remains useful in cases that do not exhibit robust overfitting.
\autoref{fig:wa_vs_no_wa_external_data} shows how the robust accuracy evolves as training progresses when using WA and additional external data (for which standard SGD does not show signs of overfitting).
We notice that the robust performance in this setting is not only preserved but even boosted when using WA. 
Hence, we formulate the hypothesis that \uline{model weight averaging helps robustness to a greater effect when robust accuracy between model iterations can be maintained}.
This hypothesis is also motivated by the observation that WA acts as a temporal ensemble -- akin to Fast Geometric Ensembling by \citealp{garipov2018loss} who argue that efficient ensembling can be obtained by aggregating multiple checkpoint parameters at different training times.
As such, it is important to ensemble a suite of equally strong and diverse models.
Although mildly successful, we note that ensembling has received some attention in the context of adversarial training~\citep{pang2019improving,strauss2017ensemble}.
In particular, \citet{tramer_ensemble_2017,grefenstette2018strength} found that ensembling could reduce the risk of gradient obfuscation caused by locally non-linear loss surfaces.

\begin{figure}[t]
\begin{center}
\centerline{\includegraphics[width=.8\columnwidth]{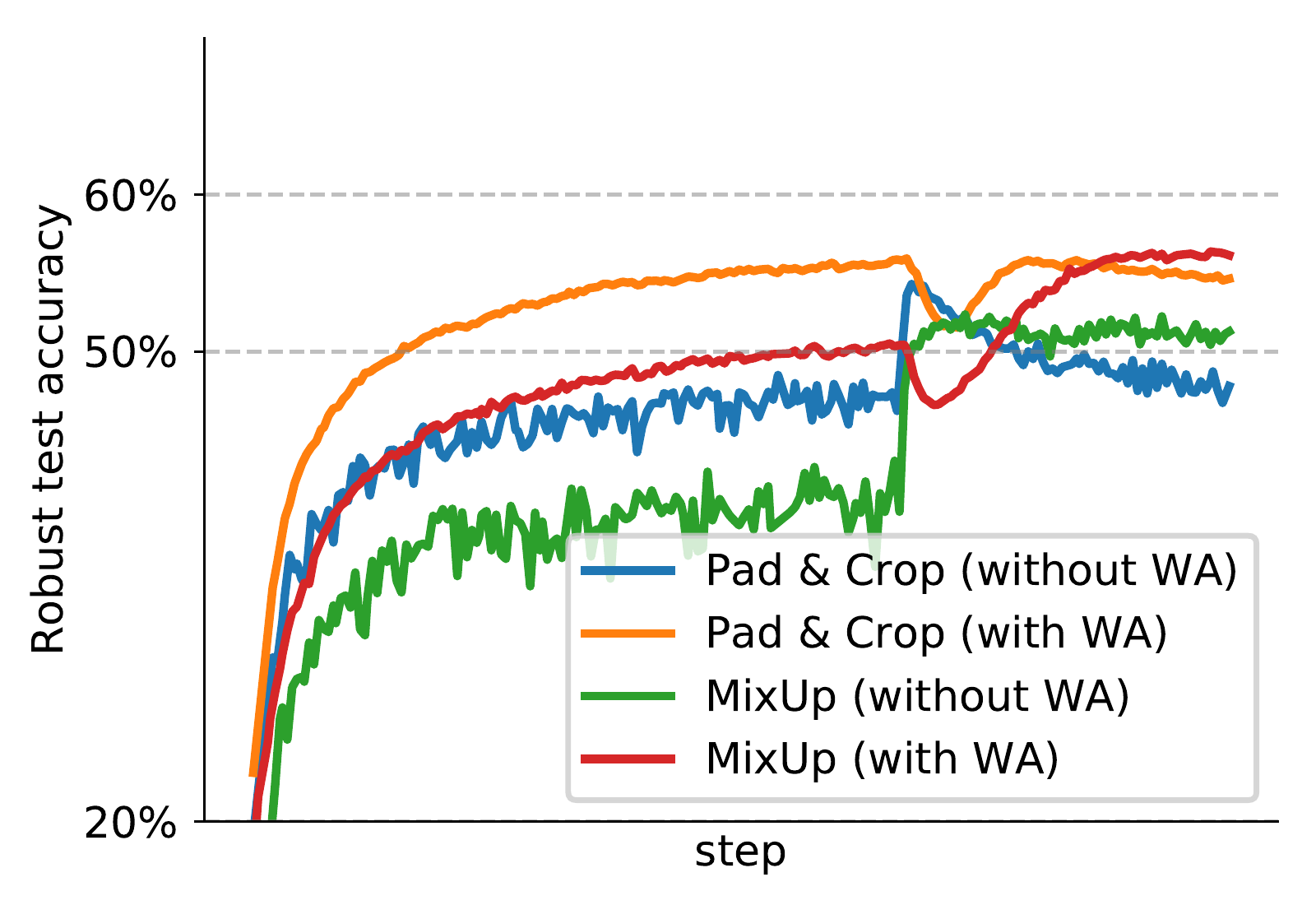}}
\caption{Accuracy against $\epsilon_\infty = 8/255$ on \cifar with and without using model weight averaging (WA) when using \emph{MixUp} or the standard combination of \emph{random padding-and-cropping} (\emph{Pad \& Crop}). The model is a \wrn-28-10 and the graph shows the robust accuracy on the test set against \pgd{40}.\label{fig:mixup_vs_padcrop_robust_verfitting}}
\end{center}
\vspace{-0.5cm}
\end{figure}

\section{Heuristics-driven Augmentations}
\label{sec:heuristics}

\paragraph{Limiting robust overfitting without external data.}

\citet{rice_overfitting_2020} show that combining regularization methods such as \emph{Cutout} or \emph{MixUp} with early stopping does not improve robustness upon early stopping alone.
While, these regularization methods do not improve upon the ``best'' robust accuracy, they reduce the extent of robust overfitting, thus resulting in a slower decrease in robust accuracy compared to classical adversarial training (which uses random crops and weight decay).
This can be seen in \autoref{fig:mixup_vs_padcrop_robust_verfitting} where \emph{MixUp} without WA exhibits no decrease in robust accuracy, whereas the robust accuracy of the standard combination of \emph{random padding-and-cropping} without WA (\emph{Pad \& Crop}) decreases immediately after the change of learning rate.

\paragraph{Verifying the hypothesis.}

Since \emph{MixUp} preserves robust accuracy, albeit at a lower level than the ``best'' obtained by \emph{Pad \& Crop}, it can be used to evaluate the hypothesis that WA is more beneficial when the performance between model iterations is maintained.
Therefore, we compare in \autoref{fig:mixup_vs_padcrop_robust_verfitting} the effect of WA on robustness when using \emph{MixUp}.
We observe that, when using WA, the performance of \emph{MixUp} surpasses the performance of \emph{Pad \& Crop}.
Indeed, the robust accuracy obtained by the averaged weights of \emph{Pad \& Crop} (in orange) slowly decreases after the change of learning rate, while the one obtained by \emph{MixUp} (in red) increases throughout training.
Ultimately, \emph{MixUp} with WA obtains a higher robust accuracy despite the fact that the non-averaged model has a significantly lower ``best'' robust accuracy than the non-averaged \emph{Pad \& Crop} model.
This finding is notable as it demonstrates for the first time the benefits of data augmentation schemes for adversarial training (this contradicts to some extent the findings from three recent publications: \citealp{rice_overfitting_2020,wu2020adversarial,gowal_uncovering_2020}).

\begin{figure*}[t]
\centering
\subfigure[Without WA]{\label{fig:augmentations_no_wa}\includegraphics[width=0.35\textwidth]{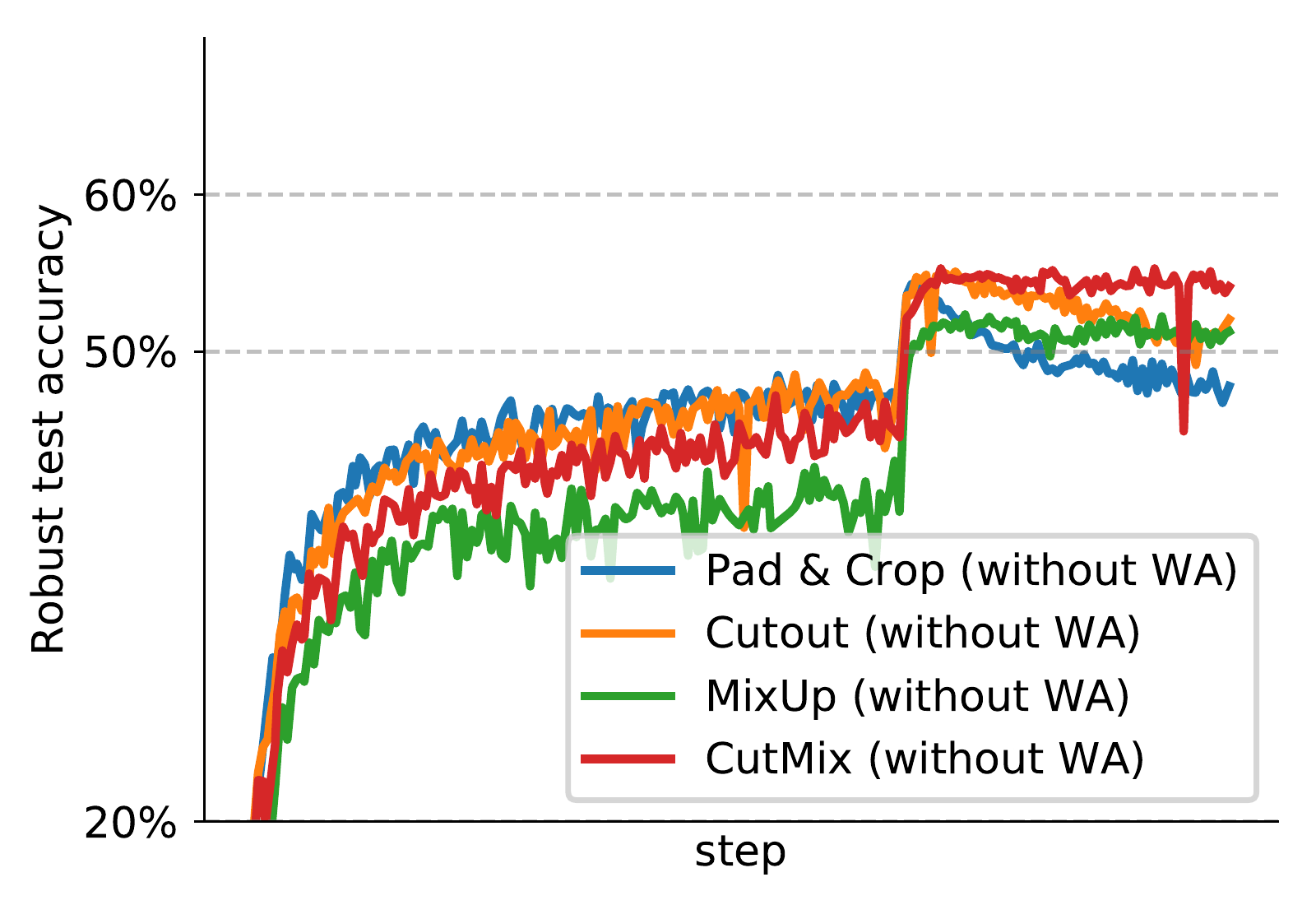}}
\hspace{1cm}
\subfigure[With WA]{\label{fig:augmentations_wa}\includegraphics[width=0.38\textwidth]{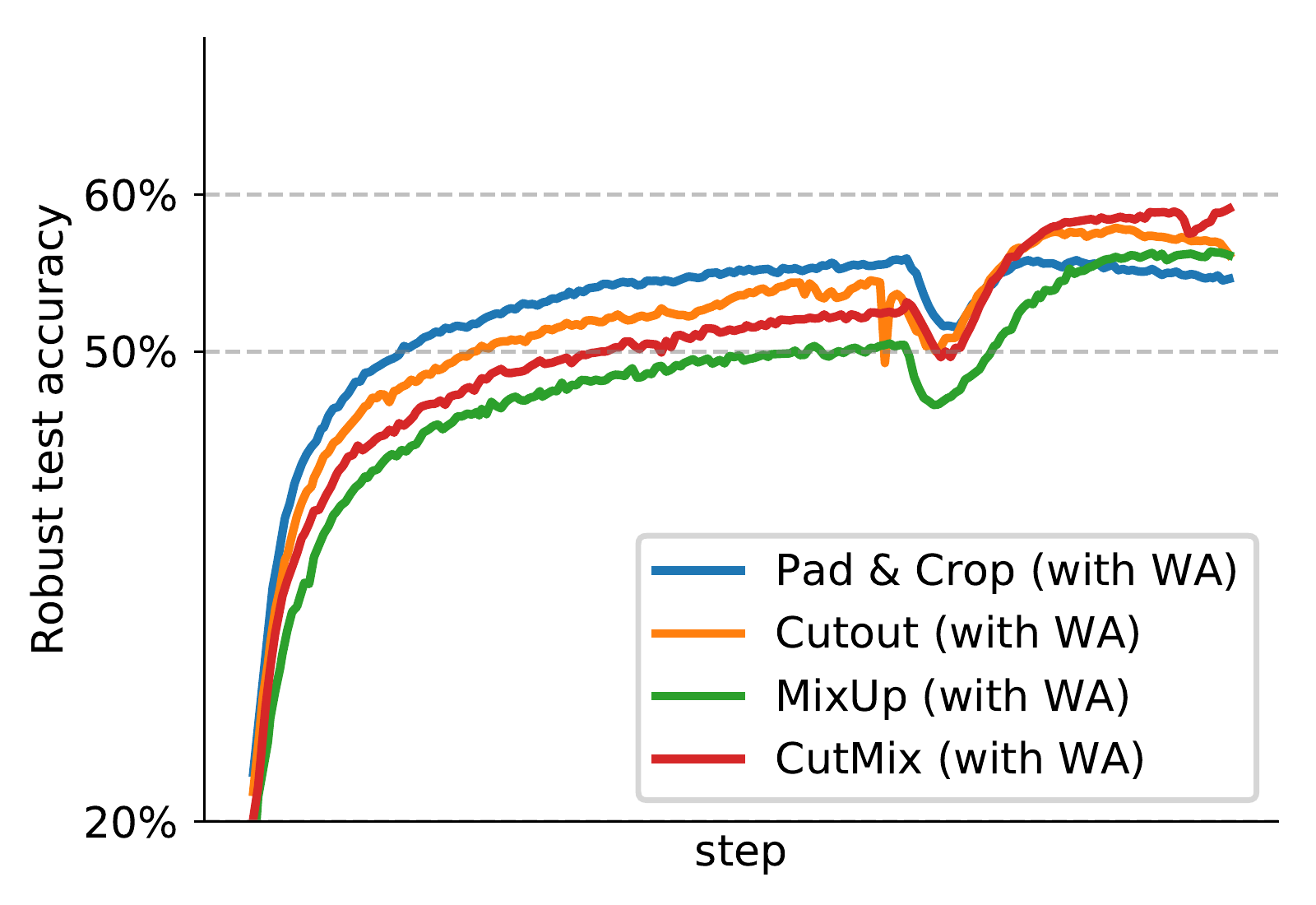}}
\caption{Accuracy against $\epsilon_\infty = 8/255$ on \cifar with and without using model weight averaging (WA) for different data augmentation schemes.
The model is a \wrn-28-10 and both panels show the evolution of the robust accuracy as training progresses (against \pgd{40}). The jump in robust accuracy two-thirds through training is due to a drop in learning rate.\label{fig:augmentations_wa_all}}
\vspace{-0.2cm}
\end{figure*}

\paragraph{Exploring heuristics-driven augmentations.}

After verifying our hypothesis for \emph{MixUp}, we investigate if other augmentations can help maintain robust accuracy and also be combined with WA to improve robustness.
We concentrate on the following image patching techniques: \emph{Cutout}~\citep{devries2017improved} which inserts empty image patches and \emph{CutMix}~\cite{yun2019cutmix} which replaces part of an image with another.
In \autoref{sec:exp_heuristics}, we also evaluate \emph{AutoAugment} \citep{cubuk_autoaugment:_2018}, \emph{RandAugment} \citep{cubuk2019randaugment} and \emph{RICAP}~\citep{takahashi2018ricap}.

We show in \autoref{fig:augmentations_wa_all} the robust accuracy obtained by these two techniques with and without WA throughout training.
All methods (including \emph{Pad \& Crop} and \emph{MixUp}) are tuned to obtain the highest robust accuracy on a separate validation set (more details in \autoref{sec:experimental_setup}).
First, we note that both \emph{Cutout} and \emph{CutMix} achieve a higher ``best'' robust accuracy than \emph{MixUp}, as shown in \autoref{fig:augmentations_no_wa}.
The ``best'' robust accuracy obtained by \emph{Cutout} and \emph{CutMix} is roughly identical to the one obtained by \emph{Pad \& Crop}, which is consistent with the results from \citet{rice_overfitting_2020}.
Second, while \emph{Cutout} suffers from robust overfitting, \emph{CutMix} does not.
Hence, as demonstrated in the previous sections, we expect WA to be more useful with \emph{CutMix}.
Indeed, we observe in \autoref{fig:augmentations_wa} that the robust accuracy of the averaged model trained with \emph{CutMix} keeps increasing throughout training and that its maximum value is significantly above the best accuracy reached by the other augmentation methods.
In \autoref{sec:exp_heuristics}, we conduct thorough evaluations of these methods against stronger attacks.

\section{Data-driven Augmentations}
\label{sec:data_driven_aug}

\paragraph{Diversity and complementarity.}

While heuristics-driven augmentations like \emph{CutMix} significantly improve robustness, there remains a large gap in robust accuracy between models trained without and with external data.
A plausible explanation is that common augmentation techniques tend to produce augmented views that are close to the original image they augment, which means that they are intrinsically limited in their ability to improve generalization.
This phenomenon is particularly exacerbated when training adversarially robust models which are known to require an amount of data polynomial in the number of input dimensions~\citep{schmidt_adversarially_2018}.
In order to improve robust generalization, it is critical to create augmentations that are more diverse and complement the training set.

In \autoref{table:similarity_train_test_self}, we sample 10K images generated by \emph{MixUp}, \emph{Cutout} and \emph{CutMix}.
For each sample, we find its closest neighbor in LPIPS~\citep{zhang2018unreasonable} feature space (more details are available in the supplementary material).
An ideal augmentation scheme would create samples that are equally likely to be close to images from the train or test sets.
We observe that neighbors of samples that were extracted from \tinyimages \footnote{Dataset from \citet{carmon_unlabeled_2019} available at \url{https://github.com/yaircarmon/semisup-adv}} are well distributed among the train and test sets, making these images particularly useful.
As expected, all three heuristics-driven augmentation schemes produce samples close to images in the train set, with \emph{CutMix} providing more diverse samples.
This observation provides an initial explanation as to why \emph{CutMix} performs better than either \emph{MixUp} or \emph{Cutout} \footnote{We note that the quality of the resulting images as well as their associated label also plays an important role.}.

\begin{table}[t]
\vspace{-0.2cm}
\caption{We sample 10K images from each heuristics-driven and each data-driven augmentation technique. For each sample in each augmented set, we find its closest neighbor in LPIPS~\citep{zhang2018unreasonable} feature space. We report the proportion of samples with a nearest neighbor in either the train set, test set or the augmented set itself (we do not match a sample with itself).
We also report the entropy (computed with the natural logarithm) of the nearest neighbor proportions (higher is better).}
\label{table:similarity_train_test_self}
\begin{center}
\resizebox{.95\columnwidth}{!}{
\begin{tabular}{l|ccc|c}
    \hline
    \cellcolor{header} \textsc{Setup} & \cellcolor{header} \textsc{Train} & \cellcolor{header} \textsc{Test} & \cellcolor{header} \textsc{Self} & \cellcolor{header} \textsc{Entropy} \TBstrut \\
    \hline
    \hline
    \multicolumn{5}{l}{\cellcolor{subheader} \textsc{Original data}} \TBstrut \\
    \hline
    \textsc{Train} & 49.81\% & 50.19\% & -- & -- \Tstrut \\
    \textsc{Test}  & 50.02\% & 49.98\% & -- & -- \\
    Extracted from \tinyimages  & 30.12\%  & 29.20\% & 40.68\% & 1.09 \Bstrut \\
    \hline
    \hline
    \multicolumn{5}{l}{\cellcolor{subheader} \textsc{Heuristics-driven Augmentations}} \TBstrut \\
    \hline
    \emph{MixUp}  & 95.93\% & 0.32\% &   3.75\% &  0.18 \Tstrut \\
    \emph{Cutout} & 94.15\% & 0.22\% &   5.63\% &  0.23 \rule{0pt}{0.0ex}\\
    \emph{CutMix}  & 84.04\% & 4.45\% &  11.51\% &  \textbf{0.53} \Bstrut \\
    \hline
    \hline
    \multicolumn{5}{l}{\cellcolor{subheader} \textsc{Data-driven Augmentations}} \TBstrut \\
    \hline
    VDVAE~\cite{child2021vdvae}  & 6.71 \%  &   5.76\%  &  87.53\%  & 0.46 \Tstrut \\
    BigGAN~\cite{brock2018large} & 11.53\%  &  10.51\%  &  77.96\%  & 0.68 \rule{0pt}{0.0ex}\\
    DDPM~\cite{ho2020denoising}   & 21.79\%  &  20.16\%  &  58.05\%  & \textbf{0.97} \Bstrut \\
    \hline
\end{tabular}
}
\vspace{-0.5cm}
\end{center}
\end{table}

\paragraph{Generative models.}

Generative models, which are capable of creating novel images, are viable augmentation candidates~\citep{dalle}.
In this work, we limit ourselves to generative models that are solely trained on the original train set, as we focus on how to improve robustness in the setting without external data.
We consider three recent and fundamentally different models: \textit{(i)} BigGAN~\citep{brock2018large}: one of the first large-scale application of \glspl{gan} which produced significant improvements in \gls{fid} and \gls{is} on \cifar (as well as on \imagenet); \textit{(ii)} \gls{vdvae}~\cite{child2021vdvae}: a hierarchical \gls{vae} which outperforms alternative \gls{vae} baselines; and \textit{(iii)} \gls{ddpm}~\cite{ho2020denoising}: a diffusion probabilistic model based on Langevin dynamics that reaches state-of-the-art \gls{fid} on \cifar.\footnote{For \gls{vdvae} and \gls{ddpm}, we use \cifar checkpoints available online (we confirmed with their authors that these checkpoints were solely trained on the \cifar train set). For BigGAN, we trained our own model and matched its \gls{is} to the one obtained by \gls{ddpm}. More details are in \autoref{sec:details_data_augment}.}
For each model, we sample 100K images per class, resulting in 1M images in total (the exact procedure is explained in \autoref{sec:details_data_augment}).
Similarly to the analysis done for heuristics-based augmentations, we sub-sample 10K images and report nearest-neighbor statistics in \autoref{table:similarity_train_test_self}.
We observe that the \gls{ddpm} distribution matches more closely the distribution from real, non-generated images extracted from \tinyimages.
We also observe that images generated by BigGAN and \gls{vdvae} tend to have their nearest neighbor among themselves which indicates that these samples are either far from the train and test distributions or produce overly similar samples.

\paragraph{Effect of data-driven augmentations on robustness.}

\begin{figure}[t]
\begin{center}
\centerline{\includegraphics[width=.8\columnwidth]{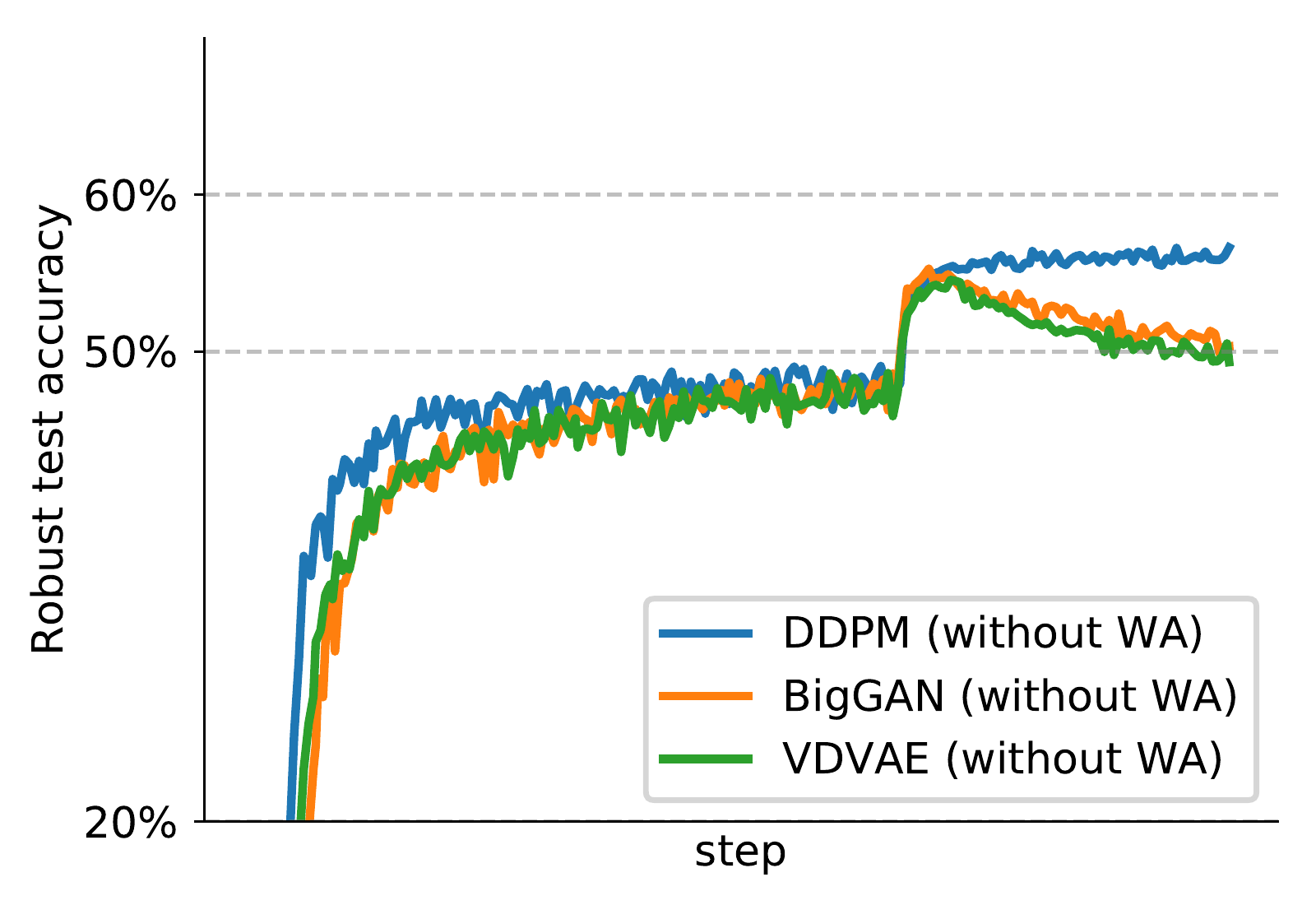}}
\caption{Accuracy against $\epsilon_\infty = 8/255$ on \cifar without using model weight averaging (WA) when adding generated samples from three different generative models. The model is a \wrn-28-10 and the graph shows the robust accuracy on the test set against \pgd{40}. The jump in robust accuracy two-thirds through training is due to a drop in learning rate.\label{fig:generated_wa_all}\vspace{-.5cm}}
\end{center}
\vspace{-1cm}
\end{figure}

While BigGAN samples lack diversity, \citet{ravuri_classification_2019} demonstrated that even these low-diversity samples could be exploited to improve the top-5 accuracy on \imagenet (albeit by a very minimal margin of 0.2\%).
An important question to ask is whether data sampled from generative models can be used to improve robustness.
We follow the procedure explained by \citet{carmon_unlabeled_2019} when training with additional external data, with the exception that we replace the external data with samples generated by each of the three generative models.
For each sample set, we tune the ratio of original-to-generated images (to be mixed for each batch) to maximize robustness on a separate validation set (in \autoref{sec:generation}).
\autoref{fig:generated_wa_all} shows how the robust accuracy evolves during training for generated sets of images without and with WA.
Only samples generated by \gls{ddpm} successfully prevent robust overfitting and result in stable performance after the drop in learning rate.
This result confirms the validity of the analysis performed in \autoref{table:similarity_train_test_self} which clearly indicates that \gls{ddpm} generates images that are closer to the test set.
See \autoref{sec:random_is_enough} for a theoretical justification that explains why generated data can improve robustness.

\vspace{-.2cm}
\section{Experimental Results}

In \autoref{sec:exp_heuristics}, we compare several heuristics-driven augmentations and evaluate their performance in various setups (model size and data settings).
Then, we explore data-driven augmentations with three popular generative models in \autoref{sec:generation}.
Finally, we posit that both approaches (data-driven and heuristics-driven) are complementary and combine them in \autoref{sec:aug_plus_gen}.
We observe that this combination improves robust accuracy to 64.20\%, compared to 57.14\% without any augmentation on \cifar against $\epsilon_\infty = 8/255$

\vspace{-.2cm}
\subsection{Experimental Setup}
\label{sec:experimental_setup}

\paragraph{Architecture.}

We use \glspl*{wrn}~\citep{he2015deep,zagoruyko2016wide} as our backbone network.
This is consistent with prior work \citep{madry_towards_2017,rice_overfitting_2020,zhang_theoretically_2019,uesato_are_2019,gowal_uncovering_2020} which use diverse variants of this network family.
Furthermore, we adopt the same architecture details as in \citet{gowal_uncovering_2020} with Swish/SiLU
~\citep{hendrycks2016gaussian} activation functions and WA~\citep{izmailov_averaging_2018}.
The decay rate of WA is set to $\tau=0.999$ and $\tau=0.995$ in the settings without and with extra data, respectively.
If not specified otherwise, the experiments are conducted on a \wrn-28-10 model which has a depth of 28 and a  width multiplier of 10 and contains 36M parameters.

\paragraph{Adversarial training.}
In all of our experiments, we use TRADES~\citep{zhang_theoretically_2019} with 10 \gls*{pgd} steps.
In the setting without added data, we train for $400$ epochs with a batch size of $512$.
When using additional generated or external data, we increase the batch size to $1024$ with a ratio of original-to-added data of 0.3 (unless stated otherwise) and train for $800$ \cifar-equivalent epochs.
More details can be found in \autoref{sec:app_setup}.

\paragraph{Evaluation.}

We follow the evaluation protocol designed by \citet{gowal_uncovering_2020}.
Specifically, we train two models for each hyperparameter setting, and pick the best model by evaluating the robust accuracy on a separate validation set of 1024 samples using \pgd{40} (we also perform early stopping on the same validation set).
Finally, for each setting, we report the robust test accuracy against a mixture of \autoattack~\citep{croce_reliable_2020} and \multitargeted~\citep{gowal_alternative_2019}, which is denoted by \textsc{AA+MT}.
As doing adversarial training with 10 \gls{pgd} steps is roughly ten times more computationally expensive than nominal training, we do not report confidence intervals.
Nevertheless, as a comparison point, we trained ten \wrn-28-10 models on \cifar with \emph{Pad \& Crop} and with \emph{CutMix}.
The resulting robust test accuracies on \cifar against $\epsilon_\infty = 8/255$ are respectively 54.44$\pm$0.39\% and 57.50$\pm$0.24\%, thus showing a relatively low variance in the results.
Furthermore, as we will see, our best models are well clear of the threshold for statistical significance.

\subsection{Heuristics-driven Augmentations}
\label{sec:exp_heuristics}

\begin{figure}[t]
\begin{center}
\includegraphics[width=.8\columnwidth]{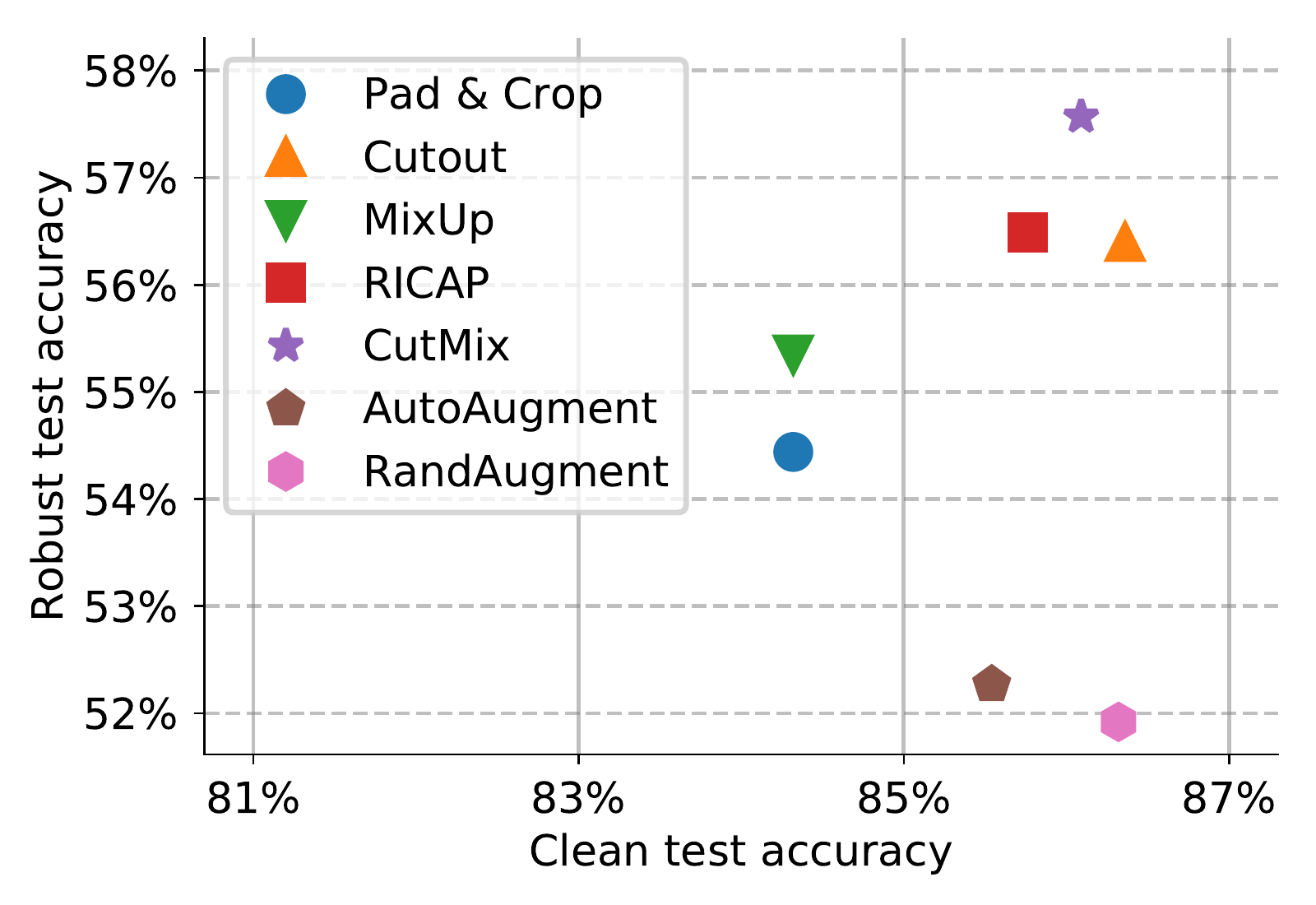}
\caption{Clean (without adversarial attacks) accuracy and robust accuracy (against \textsc{AA+MT}) for a \wrn-28-10 trained against $\epsilon_\infty = 8/255$ on \cifar for different data augmentation techniques in the setting without added data (generated or external).\label{fig:augmentations_summary}}
\end{center}
\vspace{-0.5cm}
\end{figure}

\begin{table}[t]
\caption{Robust test accuracy (against \textsc{AA+MT}) against $\epsilon_\infty = 8/255$ on \cifar as the model size increases. We compare \emph{Pad \& Crop} and \emph{CutMix}  in two data settings (with and without additional external data). Baseline numbers (from \emph{Pad \& Crop}) are slightly different from the ones obtained by \citet{gowal_uncovering_2020}, usually for the better as we use more epochs.}
\label{table:pad_vs_cutmix}
\begin{center}
\resizebox{.75\columnwidth}{!}{
\begin{tabular}{l|cc|cc}
    \hline
    \cellcolor{header} & \multicolumn{2}{c|}{\cellcolor{header} \textsc{Pad \& Crop}} & \multicolumn{2}{c}{\cellcolor{header} \textsc{CutMix}} \Tstrut \\
    \cellcolor{header} \textsc{Setup} & \cellcolor{header} \textsc{Clean} & \cellcolor{header} \textsc{Robust} & \cellcolor{header} \textsc{Clean} & \cellcolor{header} \textsc{Robust} \Bstrut \\
    \hline
    \hline
    \multicolumn{5}{l}{\cellcolor{subheader} \textsc{Without Added Data}} \TBstrut \\
    \hline
    \wrn-28-10 & 84.32\% & 54.44\% & 86.09\%  & \textbf{57.50\%} \Tstrut \\
    \wrn-34-10 & 84.89\% & 55.13\% & 86.18\%  & \textbf{58.09\%} \rule{0pt}{0.0ex} \\
    \wrn-34-20 & 85.80\% & 55.69\% & 87.80\%  & \textbf{59.25\%} \rule{0pt}{0.0ex} \\
    \wrn-70-16 & 86.02\% & 57.17\% & 87.25\%  & \textbf{60.07\%} \Bstrut \\
    \hline
    \hline
    \multicolumn{5}{l}{\cellcolor{subheader} \textsc{With 500K images from \tinyimages}} \TBstrut \\
    \hline
     \wrn-28-10 & 89.42\% & \textbf{63.05\%} & 89.90\% & 62.06\% \Tstrut \\
     \wrn-70-16 & 90.51\% & 65.88\% & 92.23\% & \textbf{66.56\%} \Bstrut \\
    \hline
\end{tabular}
}
\vspace{-0.5cm}
\end{center}
\end{table}

We consider as baseline the \emph{Pad \& Crop} augmentation which reproduces the current state-of-the-art set by \citet{gowal_uncovering_2020}.
In \autoref{fig:augmentations_summary}, we compare this baseline with various heuristics-driven augmentations, \emph{MixUp}, \emph{Cutout}, \emph{CutMix} and \emph{RICAP}, as well as learned augmentation policies with \emph{AutoAugment} and \emph{RandAugment}.
Three clusters are clearly visible.
The first cluster, containing \emph{AutoAugment} and \emph{RandAugment}, increases the clean accuracy compared to the baseline but, most notably, reduces the robust accuracy.
Indeed, these automated augmentation strategies have been tuned for standard classification, and should be adapted to the robust classification setting.
The second cluster, containing \emph{RICAP}, \emph{Cutout} and \emph{CutMix}, includes the three methods that occlude local information with patching and provide a significant boost upon the baseline with +3.06\% in robust accuracy for \emph{CutMix} and an average improvement of +1.79\% in clean accuracy.
The last cluster, with \emph{MixUp}, only improves the robust accuracy upon the baseline by a small margin of +0.91\%.
A possible explanation lies in the fact that \emph{MixUp} tends to either produce images that are far from the original data distribution (when $\alpha$ is large) or too close to the original samples (when $\alpha$ is small).
\autoref{sec:app_ablation} contains more ablation analysis on all methods.

\autoref{table:pad_vs_cutmix} shows the performance of \emph{CutMix} and the \emph{Pad \& Crop} baseline when varying the model size.
For completeness, we also evaluate our models in the setting that considers additional external data extracted from \tinyimages.
\emph{CutMix} consistently outperforms the baseline by at least +2.90\% in robust accuracy across all the model sizes in the setting without added data.
In the setting with external data, \emph{CutMix} performs worse than the baseline when the model is small (i.e., \wrn-28-10).
This is expected as the external data should generally be more useful than the augmented data and the model capacity is too low to take advantage of all this additional data.
However, \emph{CutMix} is beneficial in the setting with external data when the model is large (i.e., \wrn-70-16).
It improves upon the current state-of-the-art set by \citet{gowal_uncovering_2020} by +0.68\% in robust accuracy, thus leading to a new state-of-the-art robust accuracy of 66.56\% when using external data.

\begin{figure}[t]
\begin{center}
\includegraphics[width=.95\columnwidth]{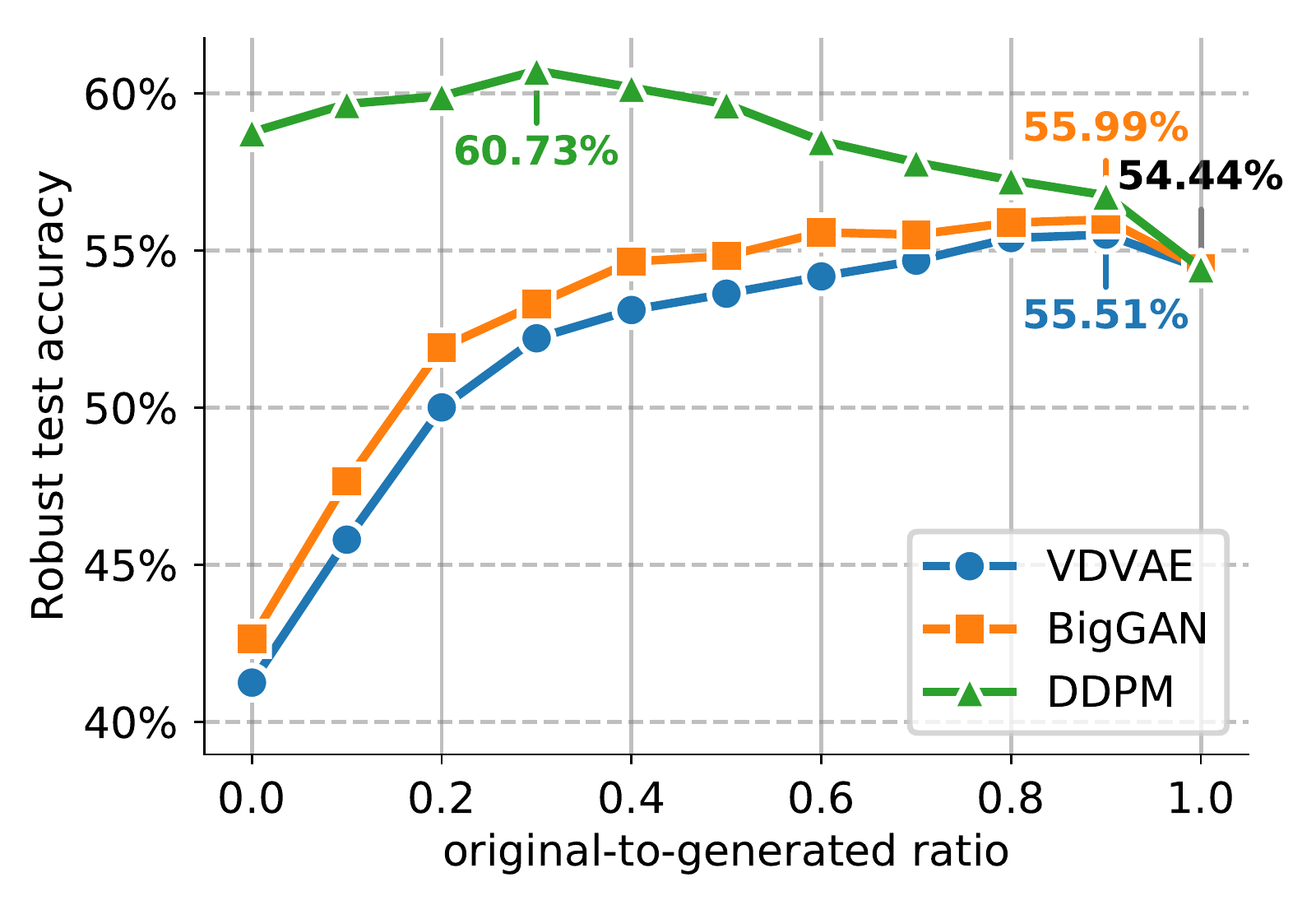}
\vspace*{-0.6cm}
\caption{Robust test accuracy (against \textsc{AA+MT}) when training a \wrn-28-10 against $\epsilon_\infty = 8/255$ on \cifar when using additional data produced by different generative models. We compare how the ratio between original images and generated images in the training minibatches affects the test robust performance (0 means generated samples only, while 1 means original \cifar train set only). We note that a \wrn-70-16 trained on \gls{ddpm} samples reaches 63.58\% robust accuracy against \textsc{AA+MT}. \label{fig:generative_ratio} \vspace{-0.cm}}
\end{center}
\end{figure}

\subsection{Data-driven Augmentations}
\label{sec:generation}

We discussed in \autoref{sec:data_driven_aug} that generative models can improve adversarial robustness by complementing the train set with additional images.
As we adversarially train the network with both 1M generated and 50K original images, we study how the robust accuracy is impacted by the mixing ratio between original and generated images in the training minibatches.
\autoref{fig:generative_ratio} explores a wide range of original-to-generated ratios for the different generative models (e.g., a ratio of 0.3 indicates that for every 3 original images, we include 7 generated images) while training a \wrn-28-10 against $\epsilon_\infty = 8/255$ on \cifar.
A ratio of zero indicates that only generated images are used, while a ratio of one indicates that only images from the \cifar training set are used.
Samples from all models improve robustness when mixed optimally, but only samples from the \gls{ddpm} improve robustness significantly.
It is also interesting to observe that using 1M generated images is better than using the 50K images from the original train set only.
While this may seem surprising, it can easily be explained if we assume that the \gls{ddpm} can produce many more high-quality, high-diversity images (some of which are visible in \autoref{sec:details_data_augment}) than the limited set of images present in the original data (c.f., \citealp{schmidt_adversarially_2018}).
Overall, \gls{ddpm} samples significantly boosts the robust accuracy with an improvement of +6.29\% compared to using the original training set only, whereas using BigGAN and \gls{vdvae} samples result in smaller (although significant) improvements upon the baseline with +1.55\% and +1.07\%, respectively.
Finally, we note that there remains a gap to the robust accuracy obtained from using 500K images from \tinyimages (see \autoref{table:pad_vs_cutmix}).

\subsection{Combining both Augmentations}
\label{sec:aug_plus_gen}

\autoref{table:combining_gen_aug} shows the performance of models obtained by combining \emph{CutMix} with samples generated by the \gls{ddpm} on \cifar against $\epsilon_\infty = 8/255$ and $\epsilon_2 = 128/255$.
First, we observe that individually adding \emph{CutMix} or generated data provides a significant boost in robust accuracy for both threat models with up to +6.44\% (in the \linf setting) and +3.81\% (in the \ltwo setting) when training a \wrn-70-16.
Second, we realize that both augmentation techniques are complementary and can be combined to further improve robustness.
Indeed, training a \wrn-70-16 with both \emph{CutMix} and generated samples sets a new state-of-the-art of 64.20\% robust accuracy on \cifar, which improves by +7.06\% upon the best model from \citet{gowal_uncovering_2020} (in the setting without external data).
We also note that combining both augmentations requires a larger model to be most beneficial (at least for the \linf setting).
Finally and most remarkably, we highlight that, despite not using any external data, our best models beat all RobustBench \footnote{\url{https://robustbench.github.io/}~\citep{croce2020robustbench}} entries that used external data (except for one).\footnote{Our models and generated data will be available at \url{https://github.com/deepmind/deepmind-research/tree/master/adversarial_robustness}}
For results on \cifarh and \svhn, please refer to \autoref{sec:app_ablation}.

\begin{table}[t]
\vspace{-0.75em}
\caption{Clean (without adversarial attacks) accuracy and robust accuracy (against \textsc{AA+MT} and \autoattack) on \cifar obtained by different models and setups in the setting with no external data.
We both test against $\epsilon_\infty = 8/255$ and $\epsilon_2 = 128/255$.\vspace*{-0.6cm}}
\label{table:combining_gen_aug}
\begin{center}
\resizebox{1.\textwidth}{!}{
\begin{tabular}{l|cc|cc}
    \hline
    \cellcolor{header} & \multicolumn{2}{c|}{\cellcolor{header} \linf} & \multicolumn{2}{c}{\cellcolor{header} \ltwo} \Tstrut \\
    \cellcolor{header} \textsc{Setup} & \cellcolor{header} \textsc{Clean} & \cellcolor{header} \textsc{Robust} & \cellcolor{header} \textsc{Clean} & \cellcolor{header} \textsc{Robust} \Bstrut \\
    \hline
    \hline
    \multicolumn{5}{l}{\cellcolor{subheader} \textsc{\wrn-28-10} (if not specified)} \TBstrut \\
    \hline
    \citet{wu2020adversarial} (\wrn-34-10) & 85.36\% & 56.17\% & 88.51\% & 73.66\% \Tstrut \\
    \citet{gowal_uncovering_2020} (trained by us)  & 84.32\% & 54.44\% & 88.60\% & 72.56\% \\
    Ours (CutMix) & 86.22\% & 57.50\% & 91.35\% & 76.12\% \\
    Ours (\gls{ddpm}) & 85.97\% & \textbf{60.73\%} & 90.24\% & 77.37\% \\
    Ours (\gls{ddpm} + CutMix) & 87.33\% & \textbf{60.73\%} & 91.79\% & \textbf{78.69\%} \Bstrut \\
    \hline
    \hline
    \multicolumn{5}{l}{\cellcolor{subheader} \textsc{\wrn-70-16}} \TBstrut \\
    \hline
    \citet{gowal_uncovering_2020}  & 85.29\% & 57.14\% & 90.90\% & 74.50\% \Tstrut \\
    Ours (CutMix) & 87.25\% & 60.07\% & 92.43\% & 76.66\% \\
    Ours (\gls{ddpm}) & 86.94\% & 63.58\% & 90.83\% & 78.31\% \\
    Ours (\gls{ddpm} + CutMix) & 88.54\% & \textbf{64.20\%} & 92.41\% & \textbf{80.38\%} \Bstrut \\
    \hline
\end{tabular}
}
\end{center}
\vspace*{-0.6cm}
\end{table}

\vspace{-.3cm}
\section{Conclusion}

Contrary to previous works~\cite{rice_overfitting_2020,gowal_uncovering_2020,wu2020adversarial}, which have tried data augmentation techniques to train adversarially robust models without success, we demonstrate that combining heuristics-based data augmentations with model weight averaging can significantly improve robustness.
Motivated by this finding, we explore generative models: we posit and demonstrate that generated samples provide a greater diversity of augmentations that help densify the image manifold and allow adversarial training to go well beyond the state-of-the-art.
Our work provides novel insights into the effect of model weight averaging on robustness, which we hope can further our understanding of robustness.

\bibliography{example_paper}
\bibliographystyle{icml2021}

\clearpage
\onecolumn
\appendix

\begin{center}
  {\Large \bf ~\\Fixing Data Augmentation to Improve Adversarial Robustness \\ (Supplementary Material)}
  \vspace{1cm}
\end{center}

\section{Experimental Setup}
\label{sec:app_setup}

\paragraph{Architecture.}

We use \glspl*{wrn}~\citep{he2015deep,zagoruyko2016wide} as our backbone network.
This is consistent with prior work \citep{madry_towards_2017,rice_overfitting_2020,zhang_theoretically_2019,uesato_are_2019,gowal_uncovering_2020} which use diverse variants of this network family.
Furthermore, we adopt the same architecture details as \citet{gowal_uncovering_2020} with Swish/SiLU
~\citep{hendrycks2016gaussian} activation functions.
Most of the experiments are conducted on a \wrn-28-10 model which has a depth of 28, a width multiplier of 10 and contains 36M parameters.
To evaluate the effect of data augmentations on wider and deeper networks, we also run several experiments using \wrn-70-16, which contains 267M parameters. 

\paragraph{Outer minimization.}
We use TRADES~\citep{zhang_theoretically_2019} optimized using SGD with Nesterov momentum~\citep{polyak1964some, nesterov27method} and a global weight decay of $5 \times 10^{-4}$. 
In the setting without added data, we train for $400$ epochs with a batch size of $512$, and the learning rate is initially set to 0.1 and decayed by a factor 10 two-thirds-of-the-way through training.
When using additional generated or external data, we increase the batch size to $1024$ with a ratio of original-to-added data of 0.3 (unless stated otherwise), train for $800$ \cifar-equivalent epochs, and use a \emph{cosine} learning rate schedule~\citep{SGDR} without restarts where the initial learning rate is set to 0.1 and is decayed to 0 by the end of training (similar to \citealp{gowal_uncovering_2020}).
In all the settings, we scale the learning rates using the linear scaling
rule of \citet{goyal2017accurate} (i.e., $\textrm{effective LR} = \max(\textrm{LR} \times \textrm{batch size} / 256, \textrm{LR})$).
We also use model weight averaging (WA)~\citep{izmailov_averaging_2018}.
The decay rate of WA is set to $\tau=0.999$ and $\tau=0.995$ in the settings without and with extra data, respectively.
Finally, to use additional generated or external data with TRADES, we annotate the extra data with the pseudo-labeling technique described by~\citet{carmon_unlabeled_2019} where a separate classifier trained on clean \cifar data provides labels to the unlabeled samples.
Note that when we use the 500K images from \tinyimages, we use the annotations already provided by \citet{carmon_unlabeled_2019}.

\paragraph{Inner minimization.}
Adversarial examples are obtained by maximizing the  Kullback-Leibler  divergence between the predictions made on clean inputs and those made on adversarial inputs~\citep{zhang_theoretically_2019}.
This optimization procedure is done using the Adam optimizer \citep{kingma_adam:_2014} for 10 \gls*{pgd} steps.
We take an initial step-size of $0.1$ which is then decreased to $0.01$ after 5 steps.
Note that we did not see any significant difference between using Adam for the inner minimization (with the aforementioned settings) and using the standard \gls{pgd} formulation in \autoref{eq:bim} with step-size $\alpha = \epsilon / 4$ and $K = 10$ steps.

\paragraph{Evaluation.}

We follow the evaluation protocol designed by \citet{gowal_uncovering_2020}.
Specifically, we train two (and only two) models for each hyperparameter setting, perform early stopping for each model on a separate validation set of 1024 samples using \pgd{40} similarly to~\citet{rice_overfitting_2020} and pick the best model by evaluating the robust accuracy on the same validation set .
Finally, we report the robust test accuracy against a mixture of \autoattack~\citep{croce_reliable_2020} and \multitargeted~\citep{gowal_alternative_2019}, which is denoted by \textsc{AA+MT}.
This mixture consists in completing the following sequence of attacks: \autopgd on the cross-entropy loss with 5 restarts and 100 steps, \autopgd on the difference of logits ratio loss with 5 restarts and 100 steps and finally \multitargeted on the margin loss with 10 restarts and 200 steps.
The training curves, such as those visible in \autoref{fig:robust_overfitting}, are always computed using \gls{pgd} with 40 steps and the Adam optimizer (with step-size decayed by 10$\times$ at step 20 and 30).

\clearpage
\section{Additional Experiments}
\label{sec:app_ablation}

\paragraph{Model weight averaging decay rate.}

\begin{figure}[t]
\begin{center}
\centerline{\includegraphics[width=.4\columnwidth]{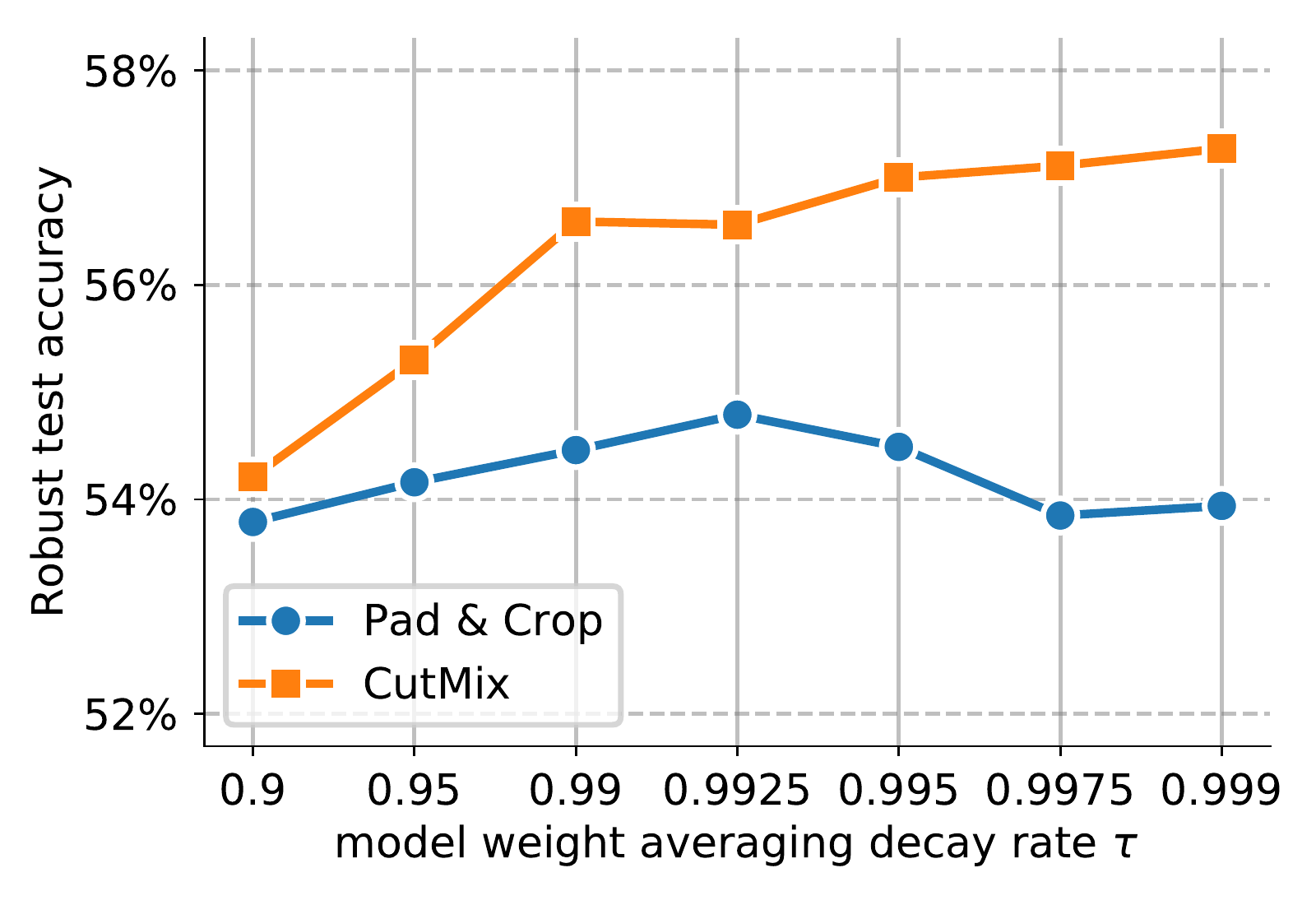}}
\caption{Robust test accuracy against AA+MT with $\epsilon_\infty= 8/255$ on \cifar as we vary the decay rate of the model weight averaging. The model is a \wrn-28-10, which is trained either with \emph{CutMix} or \emph{Pad \& Crop}.}
\label{fig:WA_decay}
\end{center}
\end{figure}

In \autoref{fig:WA_decay}, we run an ablation study measuring the robust accuracy obtained when varying the decay rate $\tau$ of model weight averaging (WA) and using either \emph{Pad \& Crop} or \emph{CutMix}.
When using \emph{CutMix}, the best robust accuracy is obtained at the highest decay rate $\tau = 0.999$.
When using \emph{Pad \& Crop}, it is only obtained at a lower decay rate $\tau=0.9925$.
This is consistent with our observation from \autoref{sec:heuristics} that highlights how WA improves robustness to a greater extent when robust accuracy can be maintained throughout training.
As larger decay rates average over longer time spans, they should better exploit the fact that \emph{CutMix} maintains robust accuracy after the learning rate is dropped to the contrary of \emph{Pad \& Crop} (see \autoref{fig:augmentations_wa_all}).

\paragraph{Mixing rate of \emph{MixUp}.}

\begin{figure*}[b]
\centering
\subfigure[\label{fig:mixup_alpha}]{\includegraphics[width=.32\columnwidth]{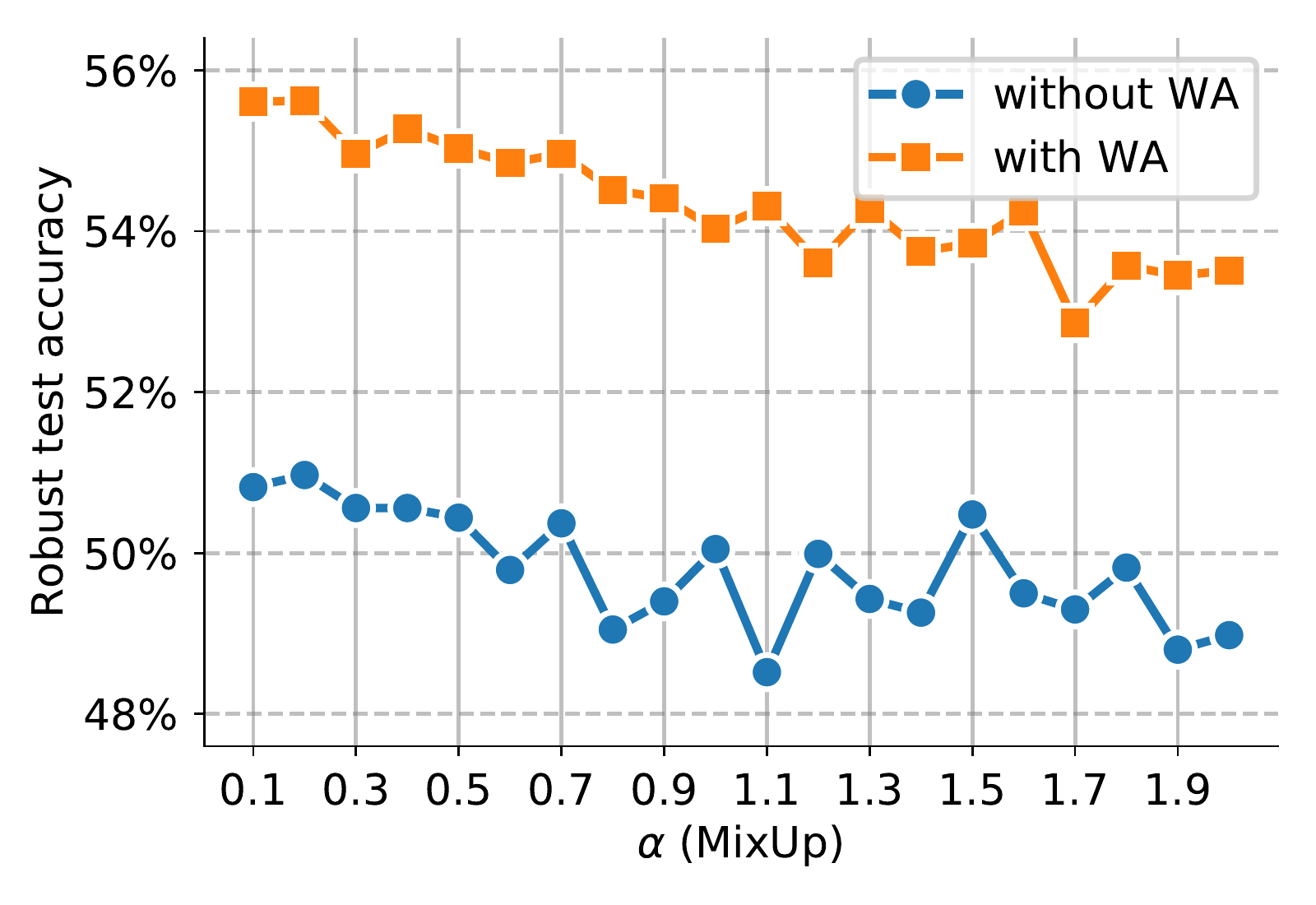}}
\subfigure[\label{fig:cutout_length}]{\includegraphics[width=.32\columnwidth]{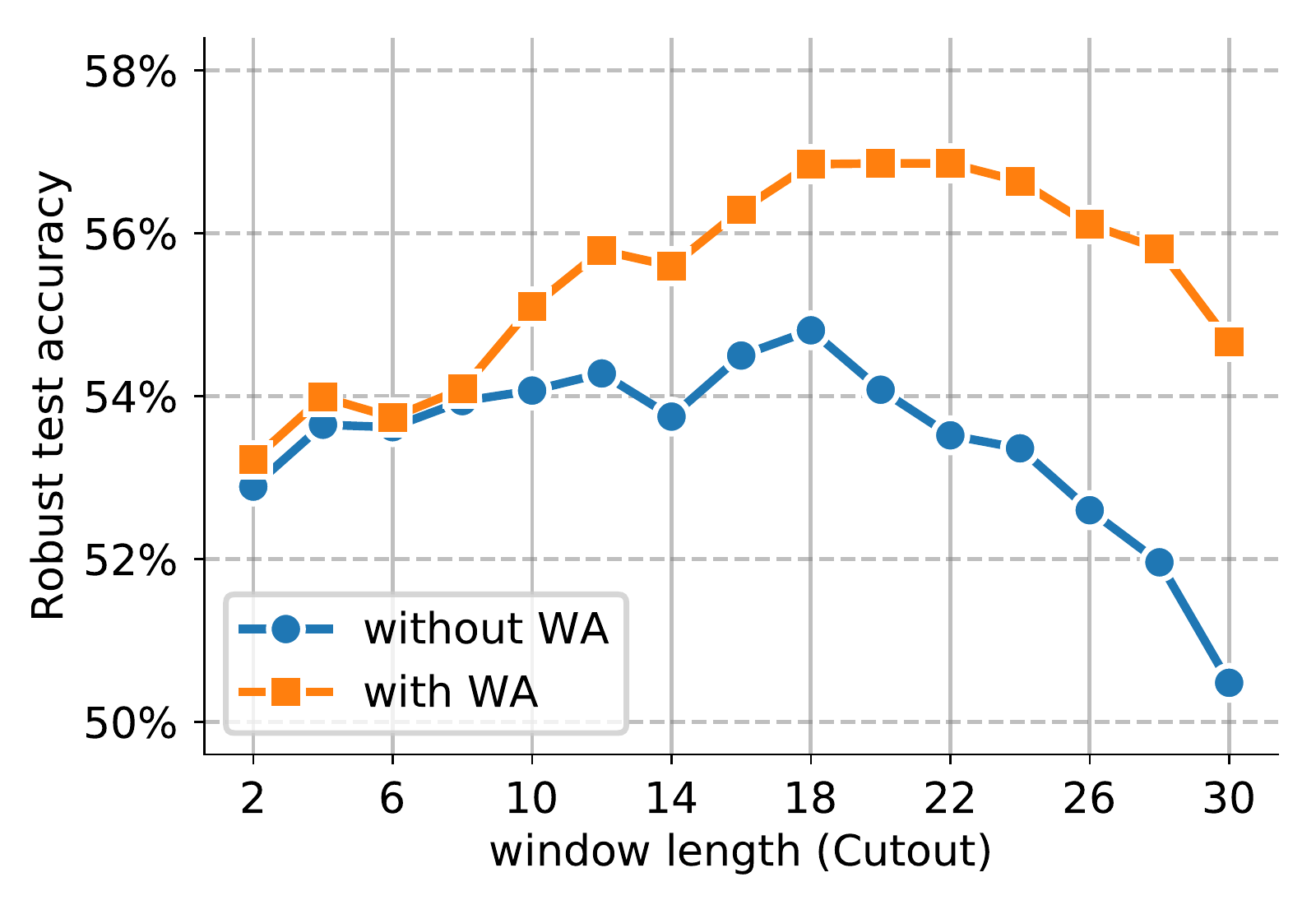}}
\subfigure[\label{fig:cutmix_length}]{\includegraphics[width=.32\columnwidth]{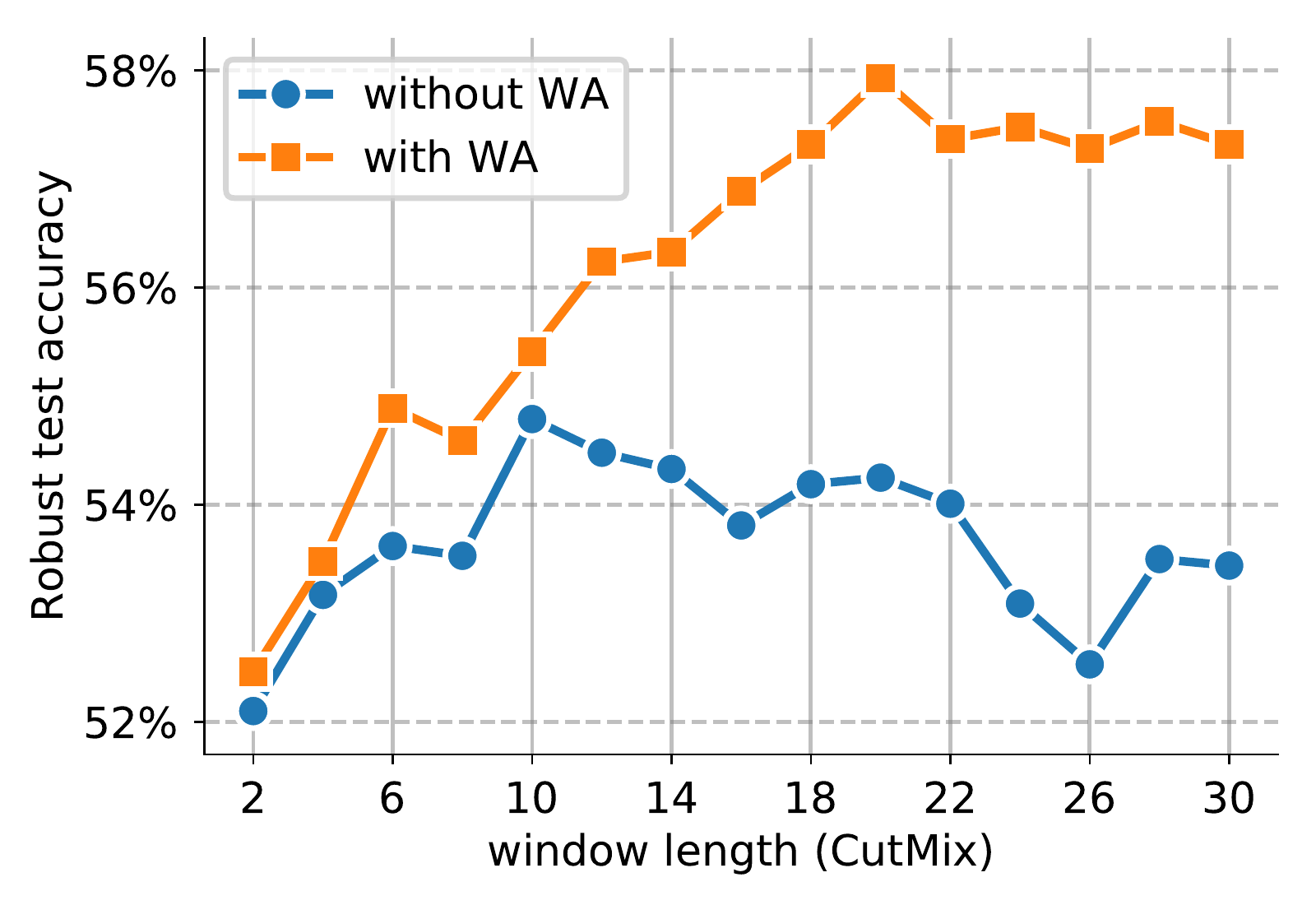}}
\caption{Robust test accuracy against AA+MT with $\epsilon_\infty= 8/255$ on \cifar as we vary \subref{fig:mixup_alpha} the mixing rate $\alpha$ of \emph{MixUp}, \subref{fig:cutout_length} the window length when using \emph{CutOut} and \subref{fig:cutmix_length} the window length when using \emph{CutMix}.  The model is a \wrn-28-10 and we compare the settings without and with model weight averaging (in which case, we use $\tau=0.999$). As a reference, the same model trained with \emph{Pad \& Crop} and model weight averaging reaches 54.44\% robust accuracy.}
\label{fig:ablation_sweeps}
\end{figure*}

For completeness, we also vary the different hyper-parameters that define the different heuristics-driven augmentations.
In particular, for \emph{MixUp}, we vary the mixing rate $\alpha$.
Remember that \emph{MixUp} blends images by sampling an interpolation point $\lambda \sim \beta(\alpha, \alpha)$ from a Beta distribution with both its parameters set to $\alpha$.
Small values of $\alpha$ produce images near the original images, while larger values tend to blend images equally.
In \autoref{fig:mixup_alpha}, we observe that smaller values of $\alpha$ are preferential (irrespective of whether we use model weight averaging).
This conclusion is in line with the recommended settings from \citet{zhang2017mixup} for standard training, but contradicts the experiments made by \citet{rice_overfitting_2020} who recommend a value of $\alpha = 1.4$ for robust training.
We also note that using model weight averaging can increase robust accuracy by up to +5.79\% when using \emph{MixUp}.

\paragraph{Window length of \emph{Cutout}.}

\emph{CutOut} creates random occlusions (i.e., anywhere in the original image) of a fixed size (measured in pixels).
Remember that \cifar images have a size of $32\times32$ pixels.
The size of this occlusion is controlled by a parameter called the \emph{window length}.
\autoref{fig:cutout_length} shows how the robust accuracy varies as a result of changing this parameter.
We notice that the optimal window length is at 18 pixels whether model weight averaging (WA) is used or not.
While WA is useful, it is noticeably less powerful when using \emph{CutOut} (as opposed to \emph{MixUp} and \emph{CutMix}) bringing only an improvement of +2.05\% in robust accuracy.
This is clearly explained by the training curves shown in \autoref{fig:augmentations_wa_all} that demonstrate that \emph{CutOut} suffers from \emph{robust overfitting}.
It also provides further evidence that support our hypothesis in \autoref{sec:hypothesis}.

\paragraph{Window length of \emph{CutMix}.}

\emph{CutMix} patches a rectangular cutout from one image onto another.
In \citet{yun2019cutmix}, the area of this patch is sampled uniformly at random (this is the setting used throughout this paper).
In this ablation experiment, however, we fix its size (i.e., window length) and observe its effect on robustness.
In \autoref{fig:cutmix_length}, we observe that the optimal size is not the same depending on whether model weight averaging (WA) is used.
We also note that WA improves robust accuracy by +3.14\%.
Overall, \emph{CutMix} obtains the highest robust accuracy of any of the four considered augmentations (including \emph{MixUp}, \emph{CutOut} and \emph{Pad \& Crop}).

\paragraph{\cifarh.}

Finally, to evaluate the generality of our approach, we evaluate \emph{CutMix} and the addition of \gls{ddpm} samples on \cifarh.
We train a new \gls{ddpm} ourselves on the train set of \cifarh and sample 1M images as described in \autoref{sec:details_data_augment}.
The results are shown in \autoref{table:cifarh_results}.
Our best model reaches 34.64\% against \autoattack and improves noticeably on the state-of-the-art (in the setting that does not use any external data).
It is worth noting that the currently best known result on \cifarh against $\epsilon_\infty = 8/255$ when using external data is 36.88\% against \autoattack.
For completeness, we also report the effect of mixing different proportions of generated and original samples in \autoref{fig:cifarh_ratio} against $\epsilon_\infty = 8/255$ on a \wrn-28-10.
Similarly to \autoref{fig:generative_ratio}, we observe that additional samples generated by \gls{ddpm} are useful to improve robustness, with an absolute improvement of +2.48\% in robust accuracy.

\paragraph{\svhn.}

We also evaluate \emph{CutMix} and the addition of \gls{ddpm} samples on \svhn.
We train a new \gls{ddpm} ourselves on the train set of \svhn and sample 1M images as described in \autoref{sec:details_data_augment}.
The results are shown in \autoref{table:svhn_results}.
Our best model reaches 61.09\% against \textsc{AA+MT} and improves noticeably on the baseline.
We also highlight that \emph{CutMix} has not been designed for \svhn and as such we should expect a smaller improvement from it.
For completeness, we also report the effect of mixing different proportions of generated and original samples in \autoref{fig:svhn_ratio} against $\epsilon_\infty = 8/255$ on a \wrn-28-10.
Similarly to \autoref{fig:generative_ratio}, we observe that additional samples generated by \gls{ddpm} are useful to improve robustness, with an absolute improvement of +4.07\% in robust accuracy.

\begin{figure}[h]
\begin{floatrow}
\capbtabbox{%
\resizebox{.5\textwidth}{!}{
\begin{tabular}{l|ccc}
    \hline
    \cellcolor{header} \textsc{Model} & \cellcolor{header} \textsc{Clean} & \cellcolor{header} \textsc{AA+MT} & \cellcolor{header} \textsc{AA} \TBstrut \\
    \hline
    \citet{cui2020learnable} (\wrn-34-10) & 60.64\% & -- & 29.33\% \Tstrut \\
    \wrn-28-10 (retrained) & 59.05\% & 28.75\% & -- \\
    \wrn-28-10 (CutMix) & 62.97\% & 30.50\% & 29.80\% \\
    \wrn-28-10 (DDPM) & 59.18\% & 31.23\% & 30.81\% \\
    \wrn-28-10 (DDPM + CutMix) & 62.41\% & \textbf{32.85\%} & \textbf{32.06\%} \Bstrut \\
    \hline
    \citet{gowal_uncovering_2020} (\wrn-70-16) & 60.86\% & 30.67\% & 30.03\% \Tstrut \\
    \wrn-70-16 (retrained) & 59.65\% & 30.62\% & -- \\
    \wrn-70-16 (CutMix) & 65.76\% & 33.24\% & 32.43\% \\
    \wrn-70-16 (DDPM) & 60.46\% & 33.93\% & 33.49\% \\
    \wrn-70-16 (DDPM + CutMix) & 63.56\% & \textbf{35.28\%} & \textbf{34.64\%} \Bstrut \\
    \hline
\end{tabular}
}
}{%
  \caption{Clean (without adversarial attacks) accuracy and robust accuracy (\textsc{AA+MT}) on \cifarh against $\epsilon_\infty = 8/255$ obtained by different models. Robust accuracy against \autoattack is also reported for select models.\label{table:cifarh_results}}%
}
\capbtabbox{%
\resizebox{.4\textwidth}{!}{
\begin{tabular}{l|ccc}
    \hline
    \cellcolor{header} \textsc{Model} & \cellcolor{header} \textsc{Clean} & \cellcolor{header} \textsc{AA+MT} \TBstrut \\
    \hline
    \wrn-28-10 (Pad \& Crop) & 92.87\% & 56.83\% \Tstrut \\
    \wrn-28-10 (CutMix) & 94.52\% & 57.32\% \\
    \wrn-28-10 (DDPM) & 94.15\% & 60.90\% \\
    \wrn-28-10 (DDPM + CutMix) & 94.39\% & \textbf{61.09\%} \Bstrut \\
    \hline
\end{tabular}
}
}{%
  \caption{Clean (without adversarial attacks) accuracy and robust accuracy (\textsc{AA+MT}) on \svhn against $\epsilon_\infty = 8/255$ obtained by different models.\label{table:svhn_results}}%
}
\end{floatrow}
\end{figure}

\clearpage
\section{Analysis of Models}

In this section, we perform additional diagnostics that give us confidence that our models are not doing any form of gradient obfuscation or masking \citep{athalye_obfuscated_2018,uesato_adversarial_2018}.

\paragraph{\autoattack and robustness against black-box attacks.}

First, we report in \autoref{table:autoattack} the robust accuracy obtained by our strongest models against a diverse set of attacks.
These attacks are run as a cascade using the \autoattack library available at \url{https://github.com/fra31/auto-attack}.
The cascade is composed as follows:
\squishlist
    \item \textsc{AutoPGD-ce}, an untargeted attack using \gls{pgd} with an adaptive step on the cross-entropy loss \citep{croce_reliable_2020},
    \item \textsc{AutoPGD-t}, a targeted attack using \gls{pgd} with an adaptive step on the difference of logits ratio \citep{croce_reliable_2020},
    \item \textsc{Fab-t}, a targeted attack which minimizes the norm of adversarial perturbations \citep{croce2020minimally},
    \item \textsc{Square}, a query-efficient black-box attack \citep{andriushchenko_square_2019}.
\squishend
First, we observe that our combination of attacks, denoted \textsc{AA+MT} matches the final robust accuracy measured by \autoattack.
Second, we also notice that the black-box attack (i.e., \textsc{Square}) does not find any additional adversarial examples.
Overall, these results indicate that our empirical measurement of robustness is meaningful and that our models do not obfuscate gradients.

\begin{table}[h]
\caption{Clean (without adversarial attacks) accuracy and robust accuracy (against the different stages of \autoattack) on \cifar obtained by different models. Refer to \url{https://github.com/fra31/auto-attack} for more details.\label{table:autoattack}}
\begin{center}
\resizebox{.9\textwidth}{!}{
\begin{tabular}{l|cc|cccc|cc}
    \hline
    \cellcolor{header} \textsc{Model} & \cellcolor{header} \textsc{Norm} & \cellcolor{header} \textsc{Radius} & \cellcolor{header} \textsc{AutoPGD-ce} & \cellcolor{header} + \textsc{AutoPGD-t} & \cellcolor{header} + \textsc{Fab-t} & \cellcolor{header} + \textsc{Square} & \cellcolor{header} \textsc{Clean} & \cellcolor{header} \textsc{AA+MT} \TBstrut \\
    \hline
    \wrn-28-10 (CutMix) & \multirow{8}{*}{\linf} & \multirow{8}{*}{$\epsilon = 8/255$} & 61.01\% & 57.61\% & 57.61\% & 57.61\% & 86.22\% & 57.50\% \Tstrut \\
    \wrn-70-16 (CutMix) & & & 62.65\% & 60.07\% & 60.07\% & 60.07\% & 87.25\% & 60.07\% \\
    \wrn-28-10 (\gls{ddpm}) & & & 63.53\% & 60.73\% & 60.73\% & 60.73\% & 85.97\% & 60.73\% \\
    \wrn-70-16 (\gls{ddpm}) & & & 65.95\% & 63.62\% & 63.62\% & 63.62\% & 86.94\% & 63.58\% \\
    \wrn-28-10 (\gls{ddpm} + CutMix) & & & 63.93\% & 60.75\% & 60.75\% & 60.75\% & 87.33\% & 60.73\% \\
    \wrn-70-16 (\gls{ddpm} + CutMix) & & & 67.30\% & 64.27\% & 64.25\% & 64.25\% & 88.54\% & 64.20\% \\
    \wrn-106-16 (\gls{ddpm} + CutMix) & & & 67.63\% & 67.63\% & 67.63\% & 64.64\% & 88.50\% & 64.58\% \\
    \wrn-70-16 (\tinyimages + CutMix) & & & 69.44\% & 66.59\% & 66.59\% & 66.58\% & 92.23\% & 66.56\% \Bstrut \\
    \hline
    \wrn-28-10 (\gls{ddpm} + CutMix) & \multirow{2}{*}{\ltwo} & \multirow{2}{*}{$\epsilon = 128/255$} & 80.00\% & 78.80\% & 78.80\% & 78.80\% & 91.79\% & 78.69\% \Tstrut \\
    \wrn-70-16 (\gls{ddpm} + CutMix) & & & 81.43\% & 80.42\% & 80.42\% & 80.42\% & 92.41\% & 80.38\% \Bstrut \\
    \hline
\end{tabular}
}
\end{center}
\end{table}

\paragraph{Further analysis of gradient obfuscation.}

\begin{figure*}[b]
\centering
\subfigure[\label{fig:eps_sweep}]{\includegraphics[width=0.4\textwidth]{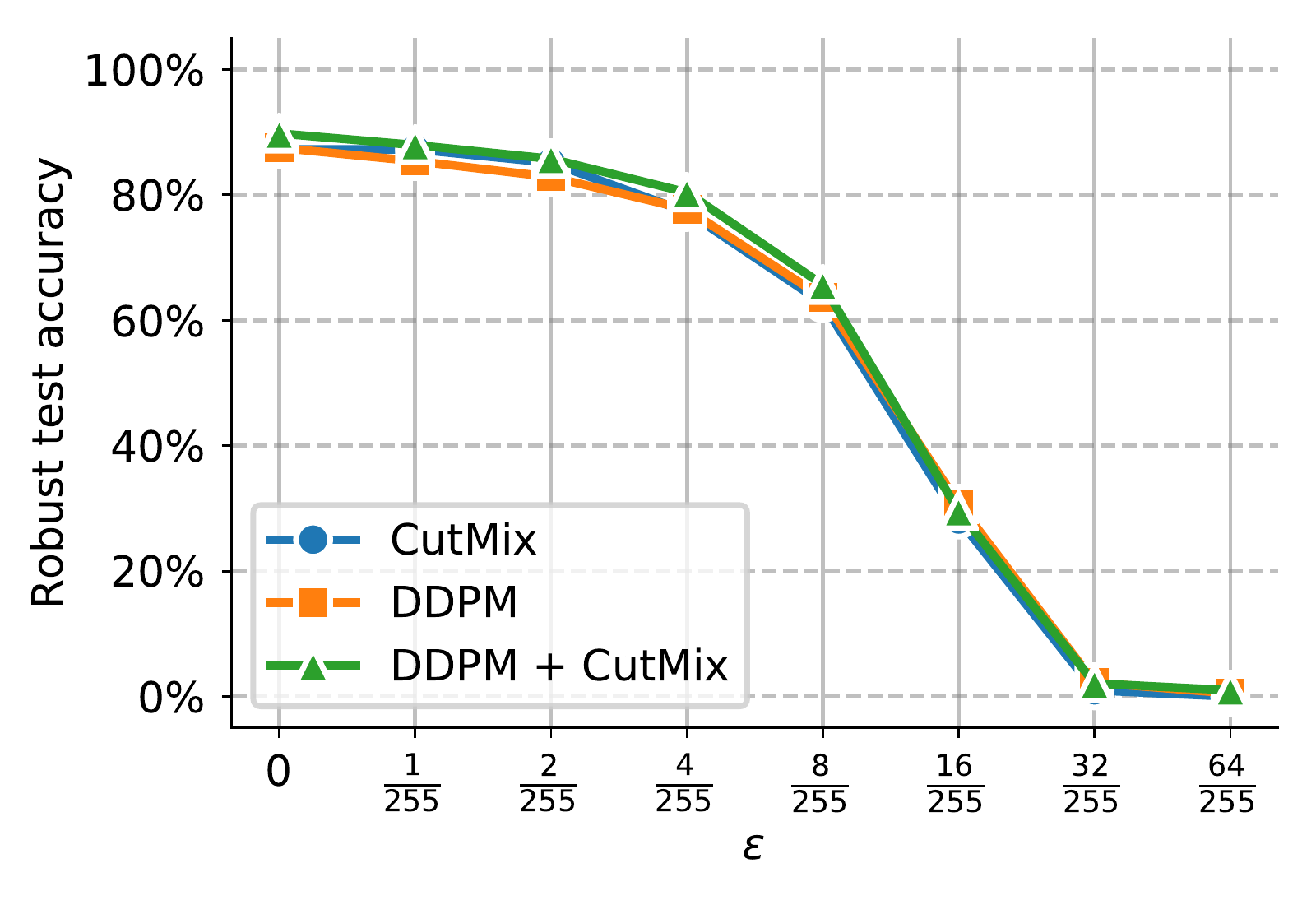}}
\subfigure[\label{fig:steps_sweep}]{\includegraphics[width=0.4\textwidth]{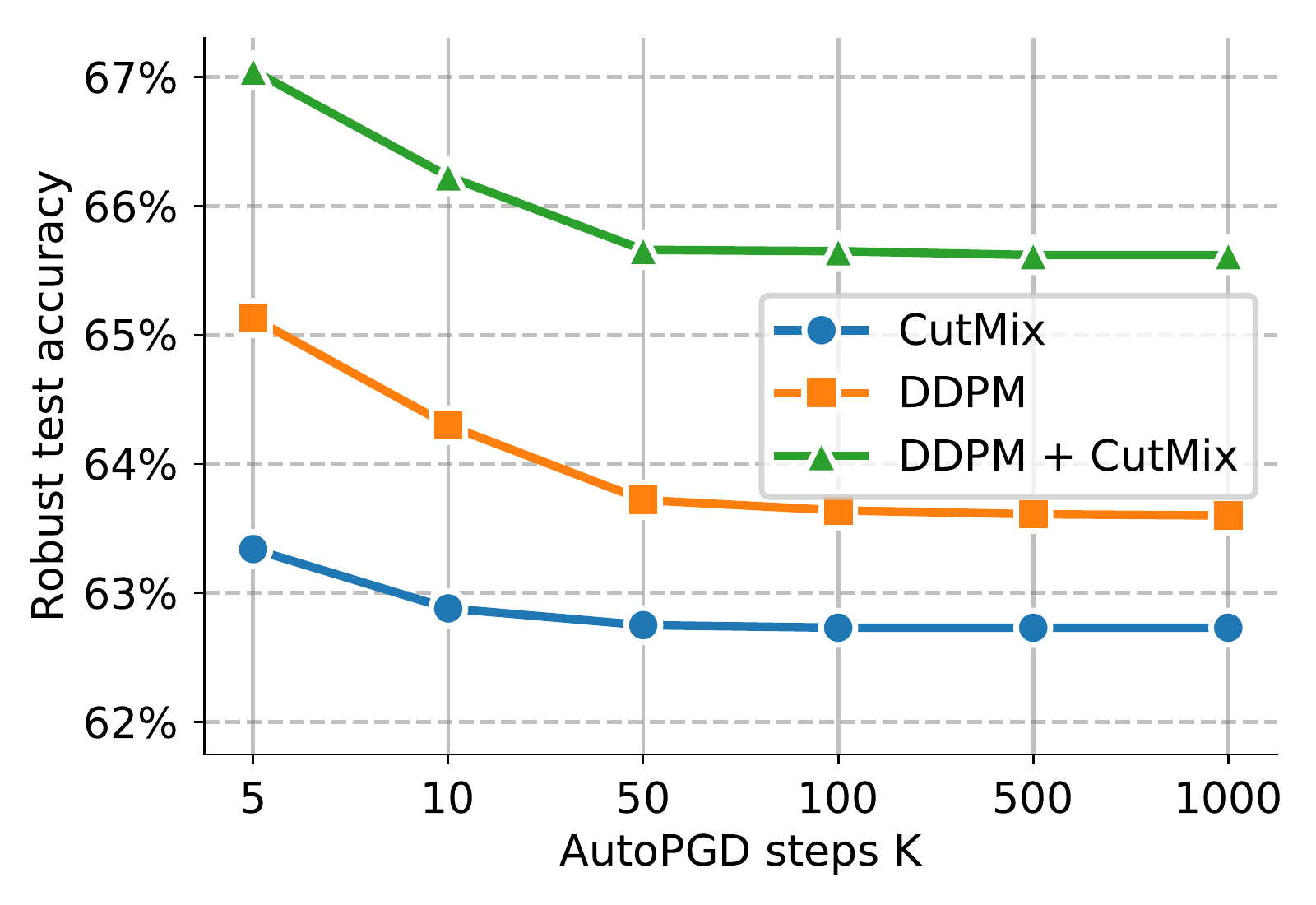}}
\caption{Robust test accuracy measured by running \textsc{AutoPGD-ce} with \subref{fig:eps_sweep} different radii $\epsilon_\infty$ and \subref{fig:steps_sweep} different number of steps $K$. Models are \wrn-70-16 networks trained with \emph{CutMix} only, \gls{ddpm} samples only and a combination of \gls{ddpm} samples and \emph{CutMix} against $\epsilon_\infty = 8/255$. They obtain 60.07\%, 61.55\% and 62.67\% robust accuracy against \textsc{AA+MT} at $\epsilon_\infty = 8/255$, respectively.}
\label{fig:grad_sweeps}
\end{figure*}

In this paragraph, we consider three \wrn-70-16 models.
These models are trained with \emph{CutMix} only, \gls{ddpm} samples only and a combination of \gls{ddpm} samples and \emph{CutMix} against $\epsilon_\infty = 8/255$. They obtain 60.07\%, 61.55\% and 62.67\% robust accuracy against \textsc{AA+MT} at $\epsilon_\infty = 8/255$, respectively.
The analysis performed here was done on an older set of models with slightly lower accuracy than our most recent models.

In \autoref{fig:eps_sweep}, we run \textsc{AutoPGD-ce} with 100 steps and 1 restart and we vary the perturbation radius $\epsilon_\infty$ between zero and $64/255$.
As expected, the robust accuracy gradually drops as the radius increases indicating that \gls{pgd}-based attacks can find adversarial examples and are not hindered by gradient obfuscation.

In \autoref{fig:steps_sweep}, we run \textsc{AutoPGD-ce} with $\epsilon_\infty = 8/255$ and 1 restart and vary the number of steps $K$ between five and 1000.
We observe that the measured robust accuracy converges after 50 steps.
This is further indication that attacks converge in $100$ steps.

\paragraph{Loss landscapes.}

Finally, we analyze the adversarial loss landscapes of the three models considered in the previous paragraph.
To generate a loss landscape, we vary the network input along the linear space defined by the worse perturbation found by \pgd{40} ($u$ direction) and a random Rademacher direction ($v$ direction).
The $u$ and $v$ axes represent the magnitude of the perturbation added in each of these directions respectively and the $z$ axis is the adversarial margin loss~\citep{carlini_towards_2017}: $z_y - \max_{i \neq y} z_i$ (i.e., a misclassification occurs when this value falls below zero).

\begin{figure*}[h]
\centering
\subfigure[\emph{CutMix} (image of a horse)]{\includegraphics[width=0.32\textwidth]{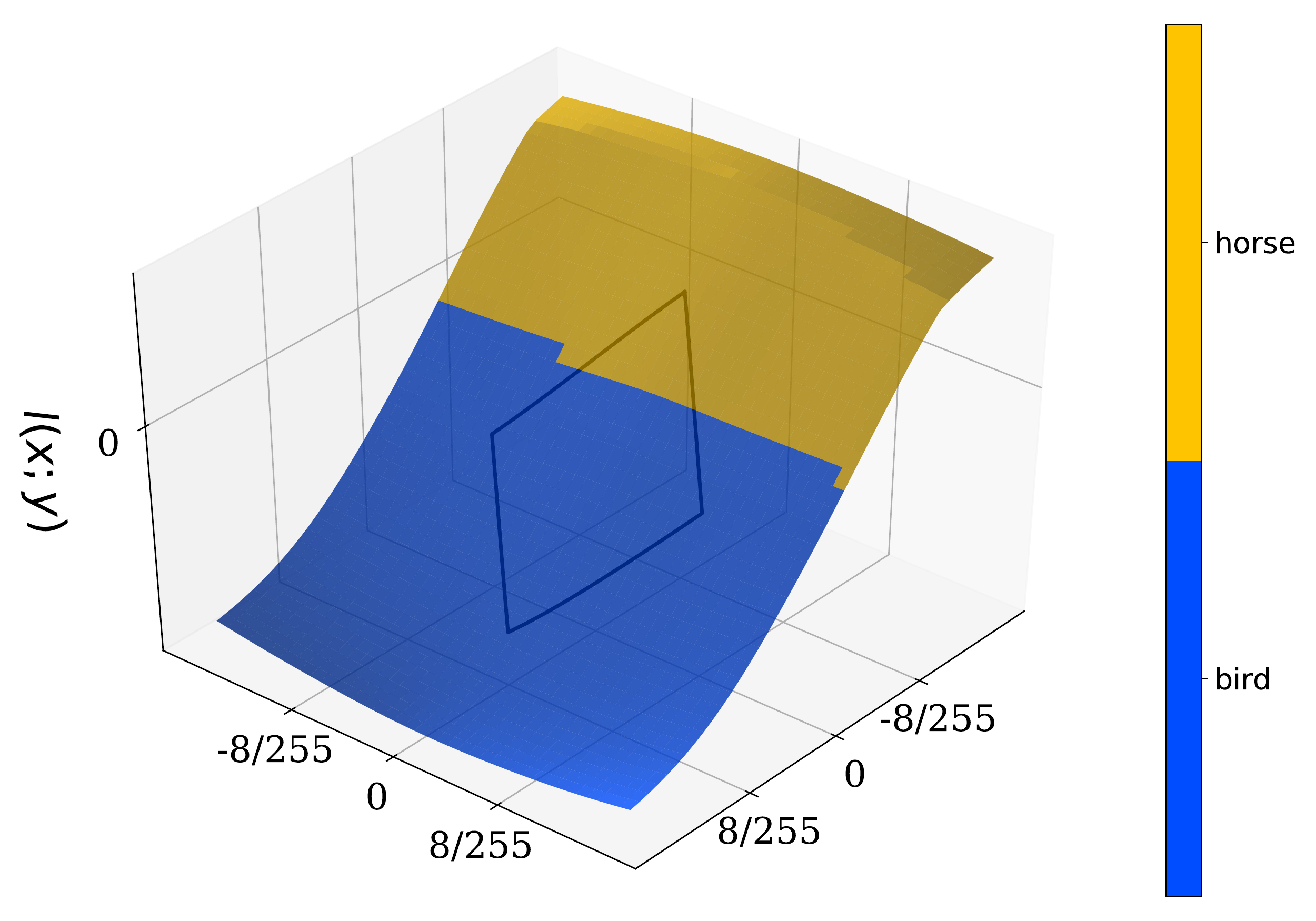}}
\subfigure[\gls{ddpm} (image of a horse)]{\includegraphics[width=0.32\textwidth]{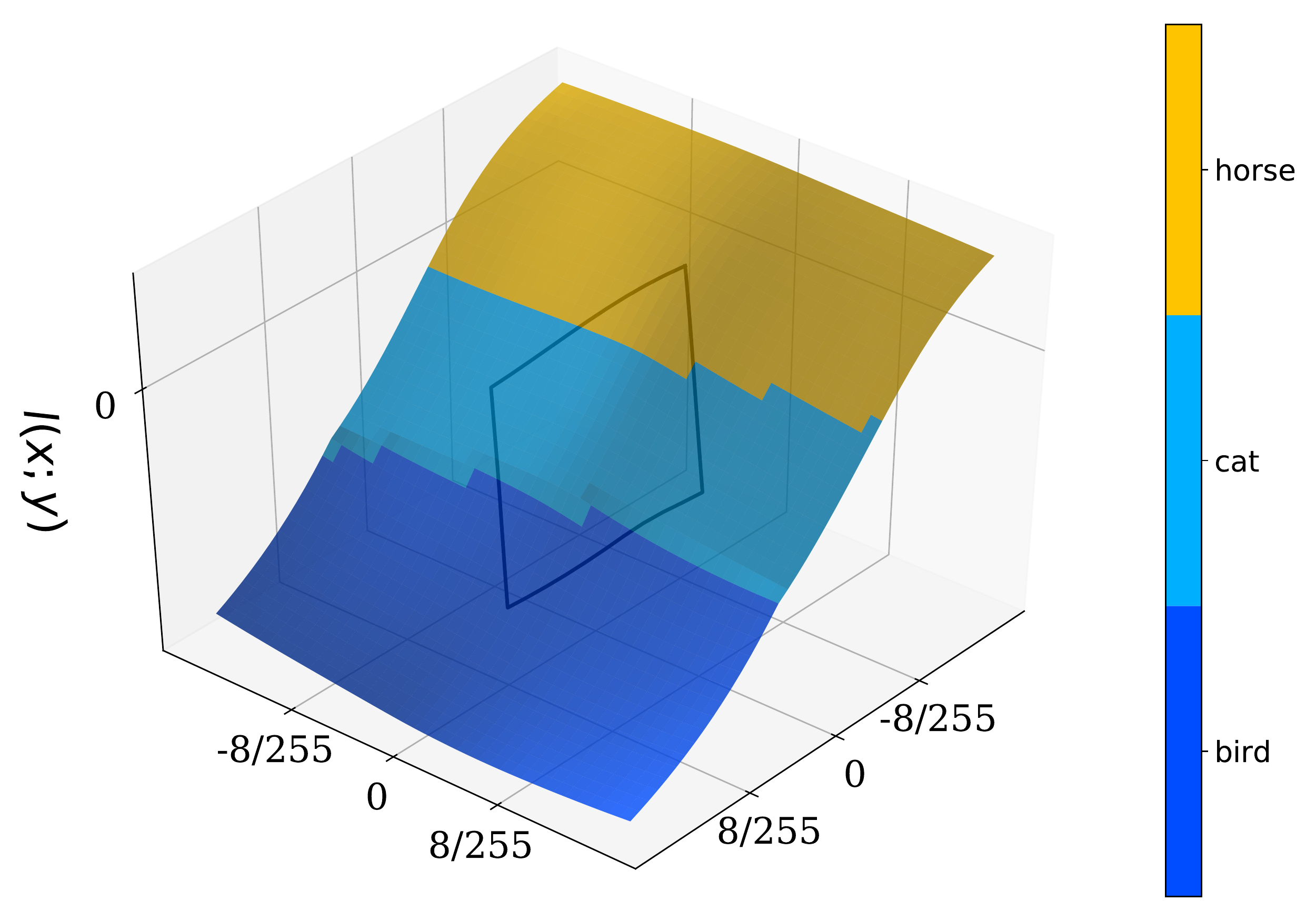}}
\subfigure[\gls{ddpm} + \emph{CutMix} (image of a horse)]{\includegraphics[width=0.32\textwidth]{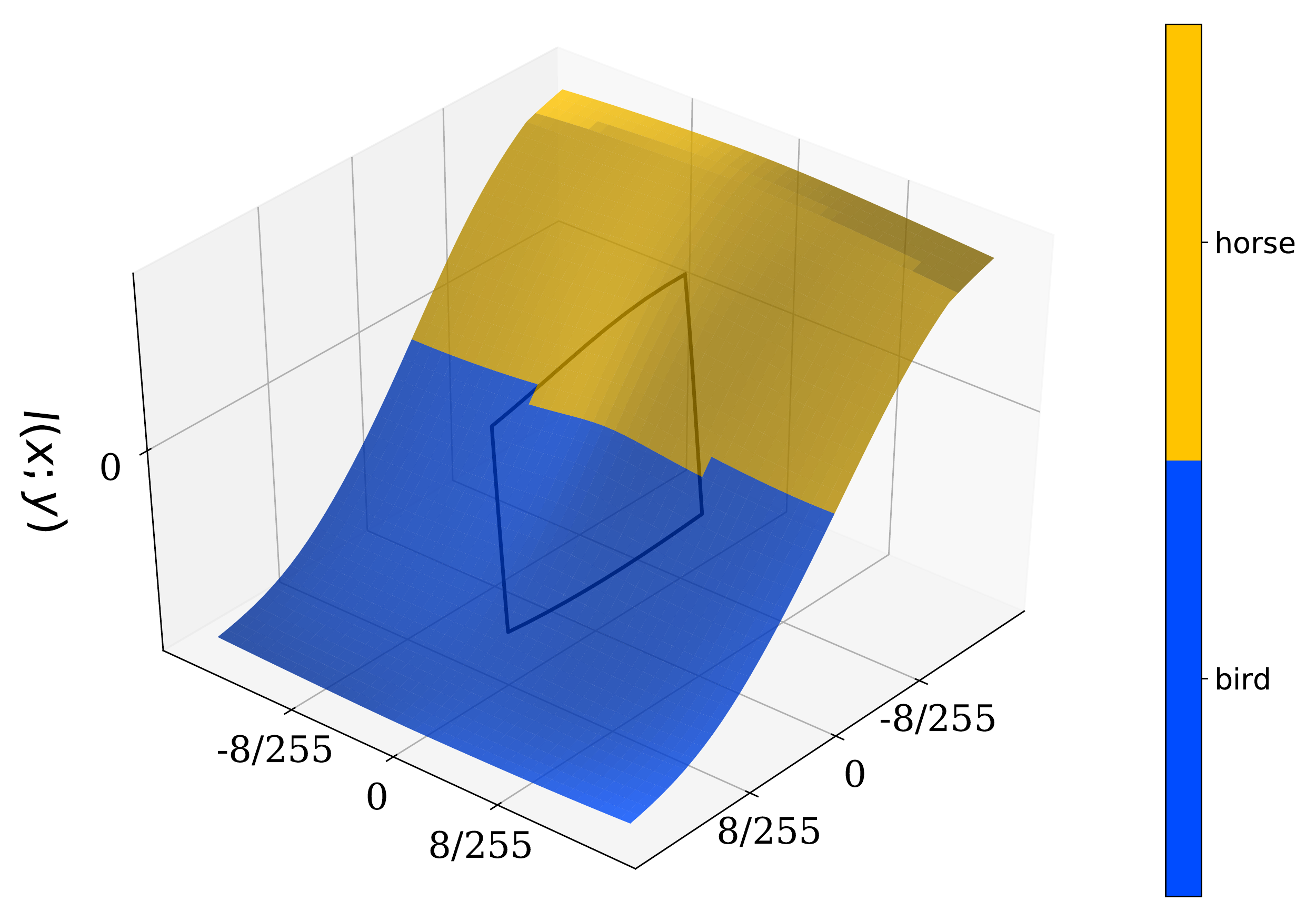}}
\subfigure[\emph{CutMix} (image of an airplane)]{\includegraphics[width=0.32\textwidth]{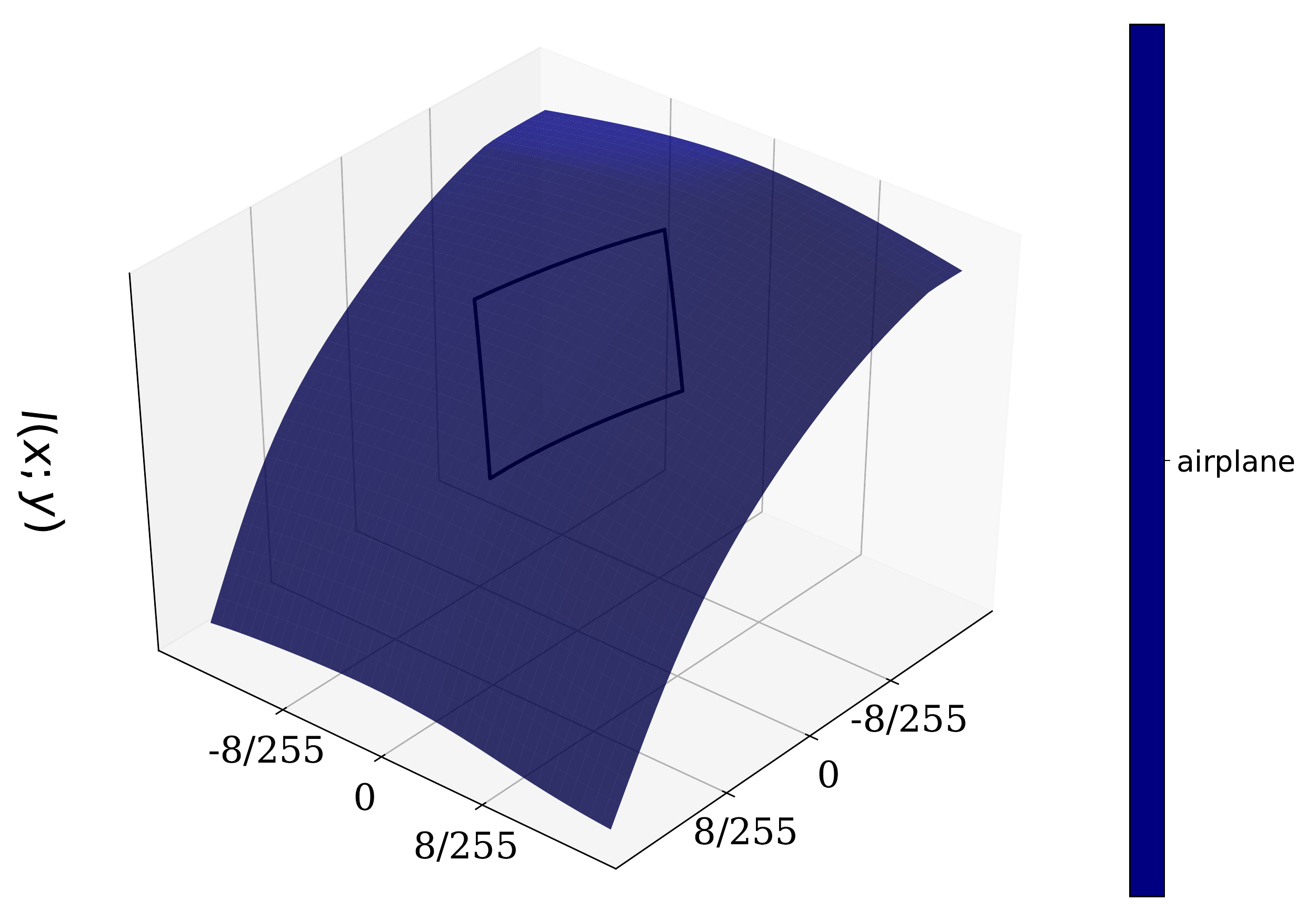}}
\subfigure[\gls{ddpm} (image of an airplane)]{\includegraphics[width=0.32\textwidth]{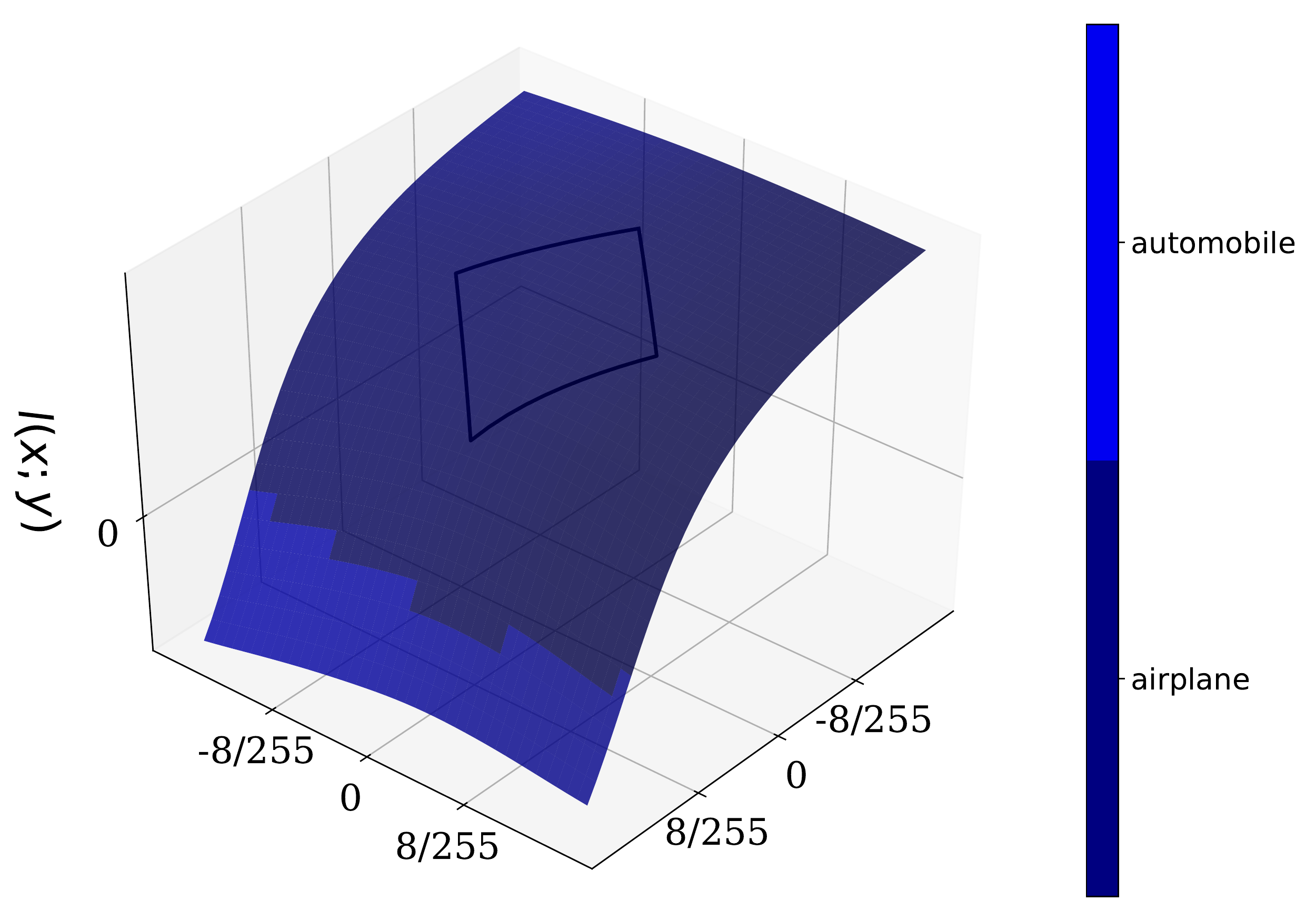}}
\subfigure[\gls{ddpm} + \emph{CutMix} (image of an airplane)]{\includegraphics[width=0.32\textwidth]{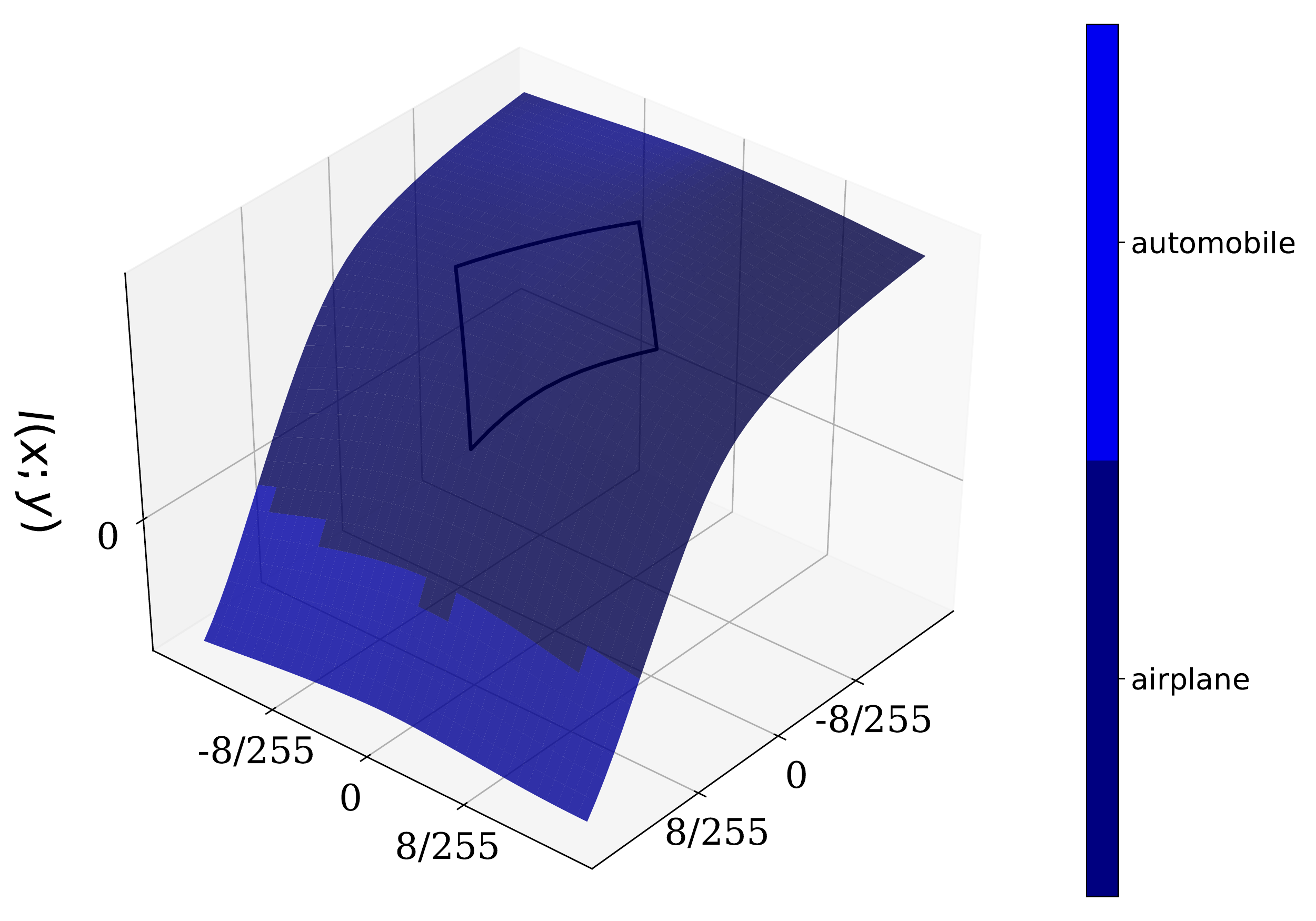}}
\caption{Loss landscapes around the first two images from the \cifar test set for the \wrn-70-16 networks trained with \emph{CutMix} only, \gls{ddpm} samples only and a combination of \gls{ddpm} samples and \emph{CutMix}. These models obtain 60.07\%, 61.55\% and 62.67\% robust accuracy, respectively.
It is generated by varying the input to the model, starting from the original input image toward either the worst attack found using \pgd{40} ($u$ direction) or a random Rademacher direction ($v$ direction). The loss used for these plots is the margin loss $z_y - \max_{i \neq y} z_i$ (i.e., a misclassification occurs when this value falls below zero). The diamond-shape represents the projected \linf ball of size $\epsilon = 8/255$ around the nominal image.}
\label{fig:linf_landscapes}
\end{figure*}

\autoref{fig:linf_landscapes} shows the loss landscapes around the first 2 images of the \cifar test set for the three aforementioned models.
All landscapes are smooth and do not exhibit patterns of gradient obfuscation.
Overall, it is difficult to interpret these figures further, but they do complement the numerical analyses done so far.

\clearpage
\section{Details on Data-driven Augmentations}
\label{sec:details_data_augment}

\paragraph{Generative models.}

In this paper, we use three different and complementary generative models: \textit{(i)} BigGAN~\citep{brock2018large}: one of the first large-scale application of \glspl{gan} which produced significant improvements in \gls{fid} and \gls{is} on \cifar (as well as on \imagenet); \textit{(ii)} \gls{vdvae}~\cite{child2021vdvae}: a hierarchical \gls{vae} which outperforms alternative \gls{vae} baselines; and \textit{(iii)} \gls{ddpm}~\cite{ho2020denoising}: a diffusion probabilistic model based on Langevin dynamics that reaches state-of-the-art \gls{fid} on \cifar.
Except for BigGAN, we use the \cifar checkpoints that are available online.
For BigGAN, we train our own model and pick the model that achieves the best \gls{fid} (the model architecture and training schedule is the same as the one used by \citealp{brock2018large}).
All models are trained solely on the \cifar train set.
As a baseline, we also fit a class-conditional multivariate Gaussian, which reaches \gls{fid} and \gls{is} metrics of 120.63 and 3.49, respectively.
We also report that BigGAN reaches an \gls{fid} of 11.07 and \gls{is} of 9.71; \gls{vdvae} reaches an \gls{fid} of 36.88 and \gls{is} of 6.03; and \gls{ddpm} reaches an \gls{fid} of 3.28 and \gls{is} of 9.44.\footnote{For \cifarh, we trained our own \gls{ddpm} which achieves an \gls{fid} of 5.58 and \gls{is} of 10.82.}

\paragraph{Datasets of generated samples.}

\begin{figure*}[h]
\centering
\subfigure[Conditional Gaussian]{\includegraphics[width=0.45\textwidth]{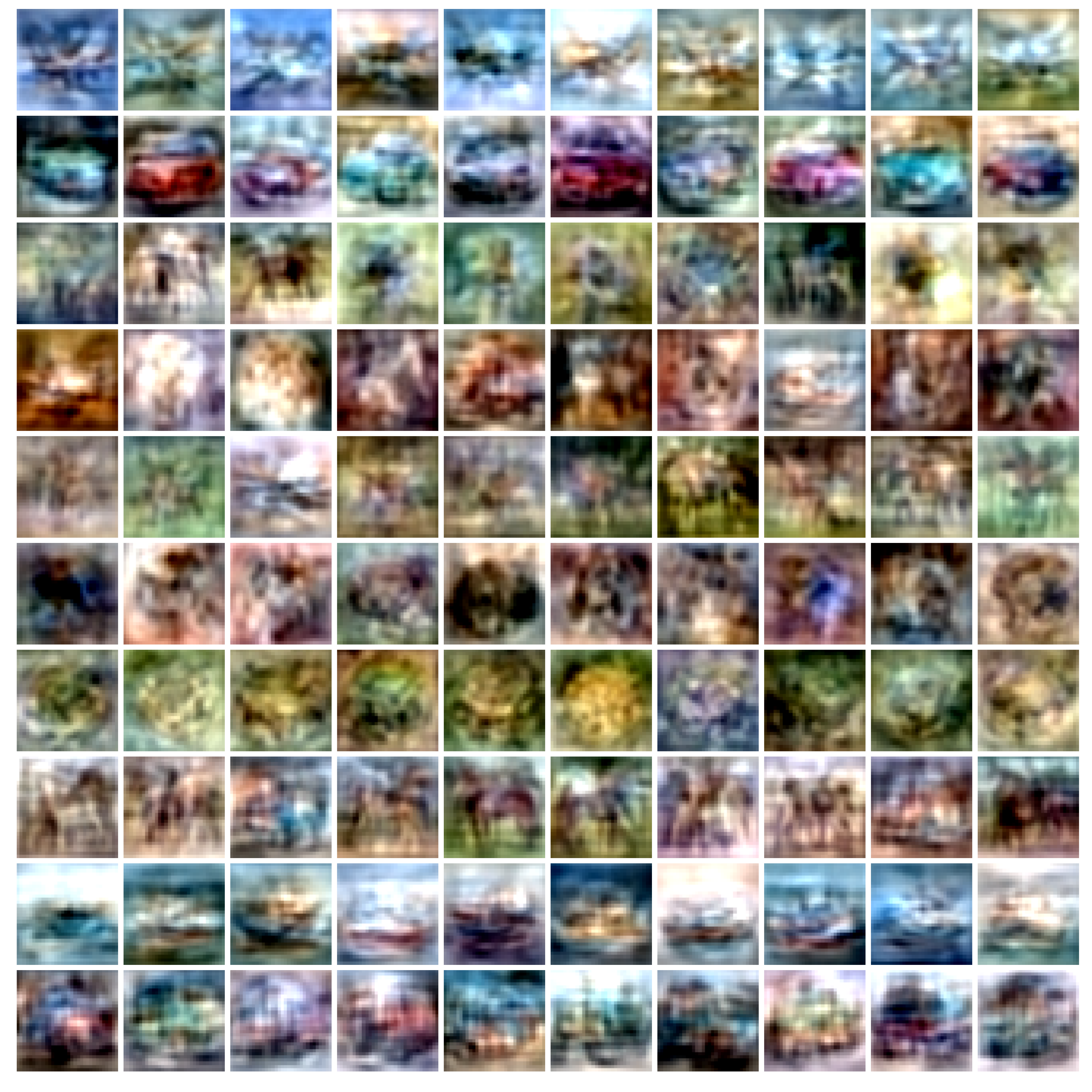}}
\subfigure[\gls{vdvae}]{\includegraphics[width=0.45\textwidth]{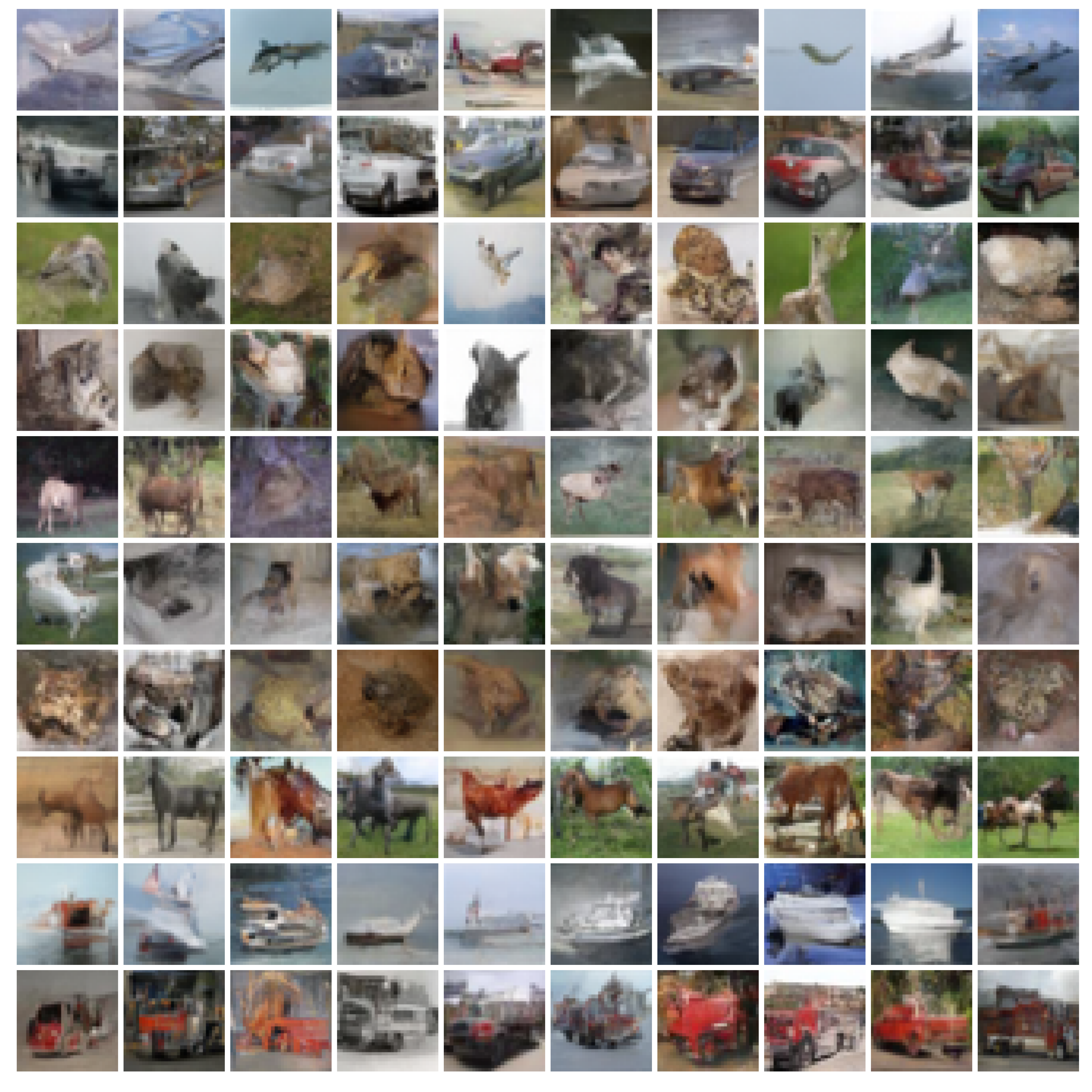}}
\subfigure[BigGAN]{\includegraphics[width=0.45\textwidth]{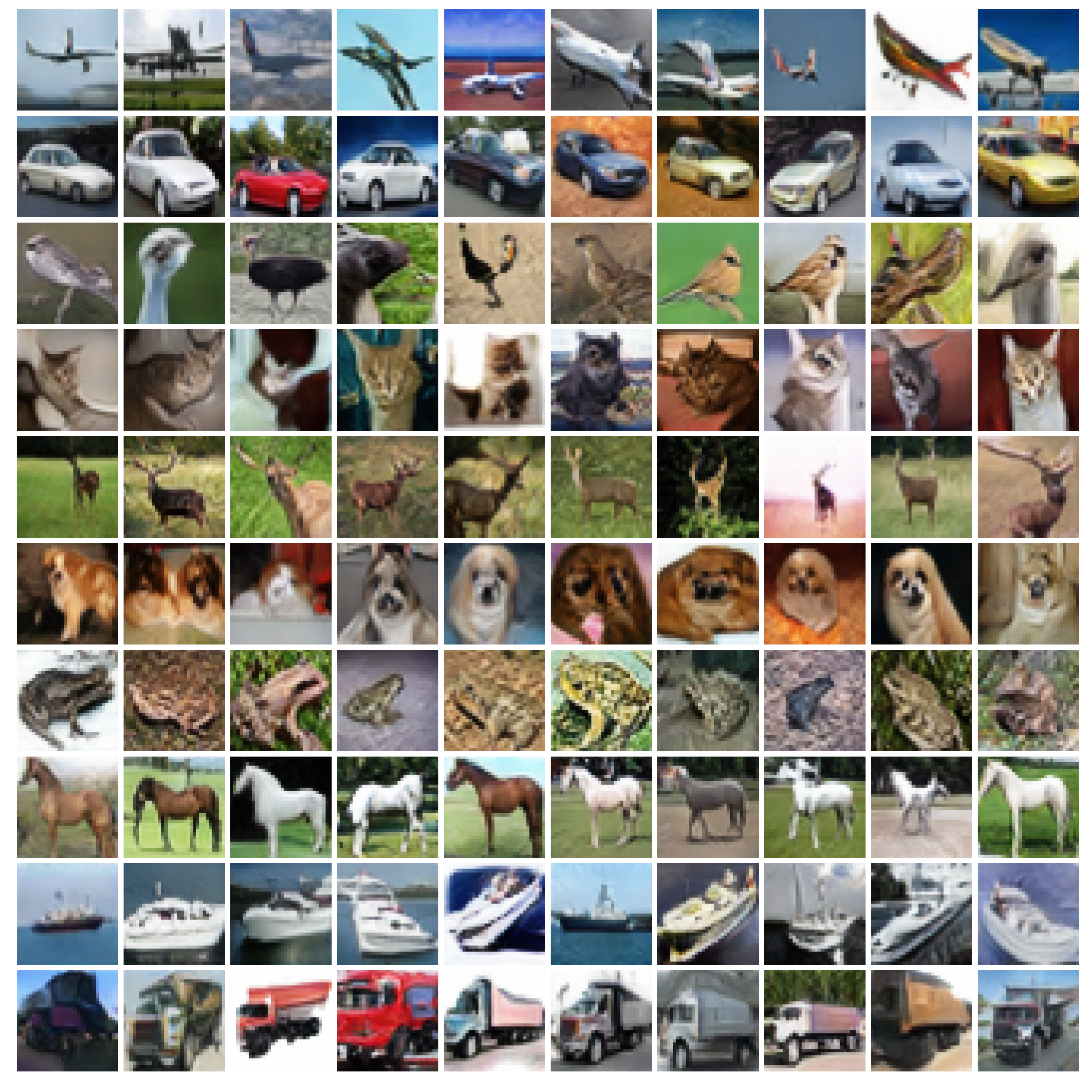}}
\subfigure[\gls{ddpm}]{\includegraphics[width=0.45\textwidth]{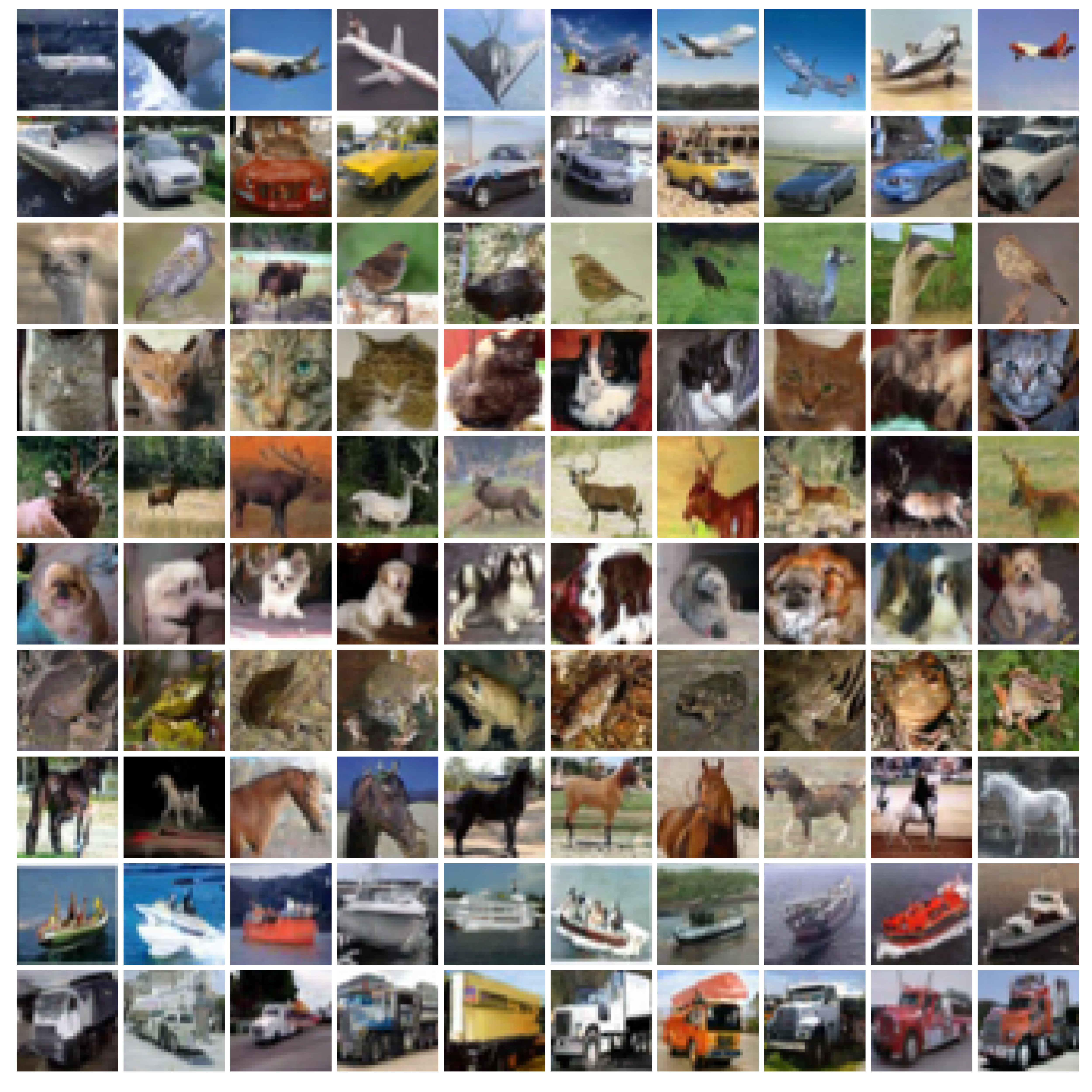}}
\caption{\cifar samples generated by different approaches and used as additional data to train adversarially robust models. Each row correspond to a different class in the following order: airplane, automobile, bird, cat, deer, dog, frog, horse, ship, truck. Each image is assigned a \emph{pseudo-label} using a standard classifier trained on the \cifar train set.}
\label{fig:samples}
\end{figure*}

We sample from each generative model 5M images.
Similarly to \citet{carmon_unlabeled_2019}, we score each image using a pretrained \wrn-28-10.
This \wrn-28-10 is trained non-robustly on the \cifar train set and achieves 95.68\% accuracy.\footnote{For \cifarh, the same model achieves 79.98\% accuracy.}
For each class, we select the top-100K scoring images and build a dataset of 1M image-label pairs.\footnote{All generated datasets will be available online at \url{https://github.com/deepmind/deepmind-research/tree/master/adversarial_robustness}.}
This additional generated data is used to train adversarially robust models by mixing for each minibatch a given proportion of original and generated examples.
\autoref{fig:samples} shows a random subset of this additional data for each generative model.
We also report the \gls{fid} and \gls{is} metrics of the resulting sets in \autoref{table:similarity_train_test_self_gen}.
They differ from the metrics obtained by each generative model as we filter images to only pick the highest scoring ones.

\paragraph{Diversity and complementarity.}

While the \gls{fid} metric does capture how two distributions of samples match, it does not necessarily provide enough information in itself to assess the overlap between the distribution of generated samples and the train or test distributions (this is especially true for samples obtained through heuristics-driven augmentations).
As such, we also decide to compute the proportion of nearest neighbors in perceptual space.
Given equal Inception metrics, a better generative model would produce samples that are equally likely to be close to training, testing or generated images.

\begin{table}[t]
\vspace{-0.2cm}
\caption{Similar to \autoref{table:similarity_train_test_self}, this table only shows generative approaches (including conditional Gaussian-fitted samples). We also add the Inception Score (IS) and Frechet Inception Distance (FID) computed from 50K samples from each set. Each set has 1M images, except for the extracted \tinyimages set which contains 500K images. Each generated set is created by randomly sampling 5M images from each generative model and taking the 100K images that score highest for each of the \cifar classes (according to a classifier trained on \cifar), which explains some difference with FID and IS numbers reported in the literature.
\label{table:similarity_train_test_self_gen}}
\begin{center}
\resizebox{.7\textwidth}{!}{
\begin{tabular}{p{0.25\textwidth}|ccc|c|cc}
    \hline
    \cellcolor{header} & \multicolumn{4}{c|}{\cellcolor{header} \textsc{Neighbor Distribution}}  & \multicolumn{2}{c}{\cellcolor{header} \textsc{Inception Metrics}} \TBstrut \\
    \cellcolor{header} \textsc{Setup} & \cellcolor{header} \textsc{Train} & \cellcolor{header} \textsc{Test} & \cellcolor{header} \textsc{Self} & \cellcolor{header} \textsc{Entropy} $\uparrow$ & \cellcolor{header} \textsc{Is} $\uparrow$ & \cellcolor{header} \textsc{Fid} $\downarrow$ \TBstrut \\
    \hline
    Extracted from \tinyimages  & 30.12\%  & 29.20\% & 40.68\% & 1.09 & $11.78 \pm 0.12$ & 2.80 \TBstrut \\
    \hline
    Conditional Gaussian  & 0.73\%  &  0.72\%  &  98.55\%  & 0.09 & $3.64 \pm 0.03$ & 117.62 \Tstrut \\
    VDVAE~\cite{child2021vdvae}  & 6.71 \%  &   5.76\%  &  87.53\%  & 0.46 & $6.88 \pm 0.05$ & 26.44 \\
    BigGAN~\cite{brock2018large} & 11.53\%  &  10.51\%  &  77.96\%  & 0.68 & \textbf{9.73}$\pm$ 0.07 & 13.78 \\
    DDPM~\cite{ho2020denoising}   & 21.79\%  &  20.16\%  &  58.05\%  & \textbf{0.97} & $9.41 \pm 0.13$ & \textbf{6.84} \Bstrut \\
    \hline
\end{tabular}
}
\vspace{-0.5cm}
\end{center}
\end{table}

We now describe how we compute \autoref{table:similarity_train_test_self} and \autoref{table:similarity_train_test_self_gen} which report the nearest-neighbors statistics for the different augmentation methods.
First, we sample 10K images from the train set of \cifar (uniformly across classes) and take the full test set of \cifar.
We then pass these 20K images through the pretrained VGG network which measures a Perceptual Image Patch Similarity, also known as LPIPS \citep{zhang2018unreasonable}.
We use the resulting concatenated activations and compute their top-100 PCA components, as this allows us to compare samples in a much lower dimensional space (i.e., 100 instead of 124,928).
Finally, for each augmentation method (heuristics- or data-driven), we sample 10K images and pass them through the pipeline composed of the LPIPS VGG network and the PCA projection computed on the original data.
For each sample, we find its closest neighbor in the PCA-reduced feature space and measure whether this nearest-neighbor belongs to the original dataset (train or test) or to the generated set (self) of 10K images.

\paragraph{Additional results.}

\begin{figure*}[h]
\centering
\subfigure[\gls{vdvae}, BigGAN and \gls{ddpm}\label{fig:generative_ratio_v2}]{\includegraphics[width=.4\columnwidth]{images/generative_ratio.pdf}}
\subfigure[\gls{ddpm} and conditional Gaussian-fit \label{fig:generative_ratio_gaussian}]{\includegraphics[width=.4\columnwidth]{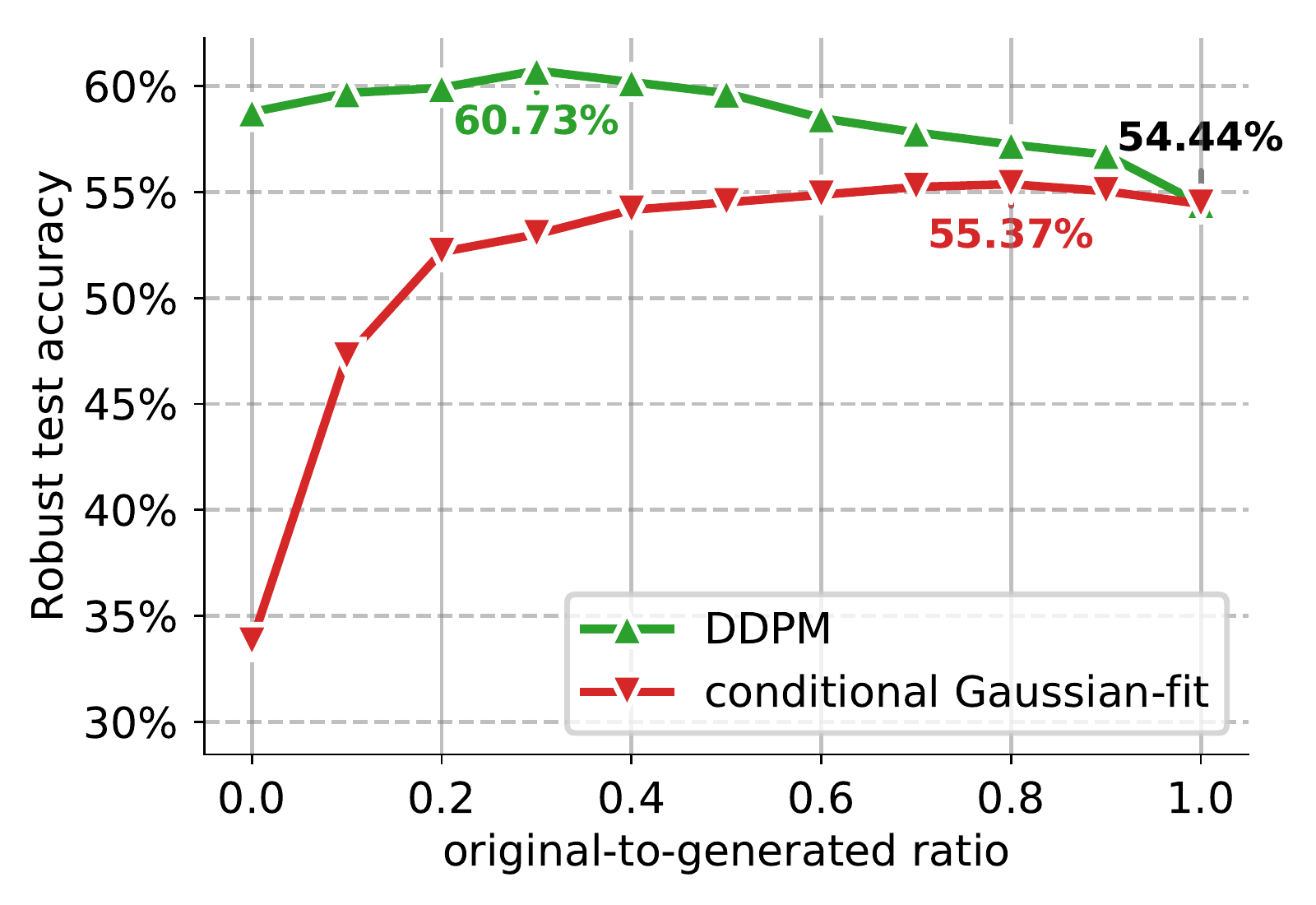}}
\caption{Robust test accuracy (against \textsc{AA+MT}) when training a \wrn-28-10 against $\epsilon_\infty = 8/255$ on \cifar when using additional data produced by different generative models. We compare how the ratio between original images and generated images in the training minibatches affects the test robust performance (0 means generated samples only, while 1 means original \cifar train set only).\label{fig:generative_ratio_v2_all}}
\end{figure*}

\begin{figure*}[h]
\centering
\subfigure[\gls{ddpm} on \cifarh\label{fig:cifarh_ratio}]{\includegraphics[width=.4\columnwidth]{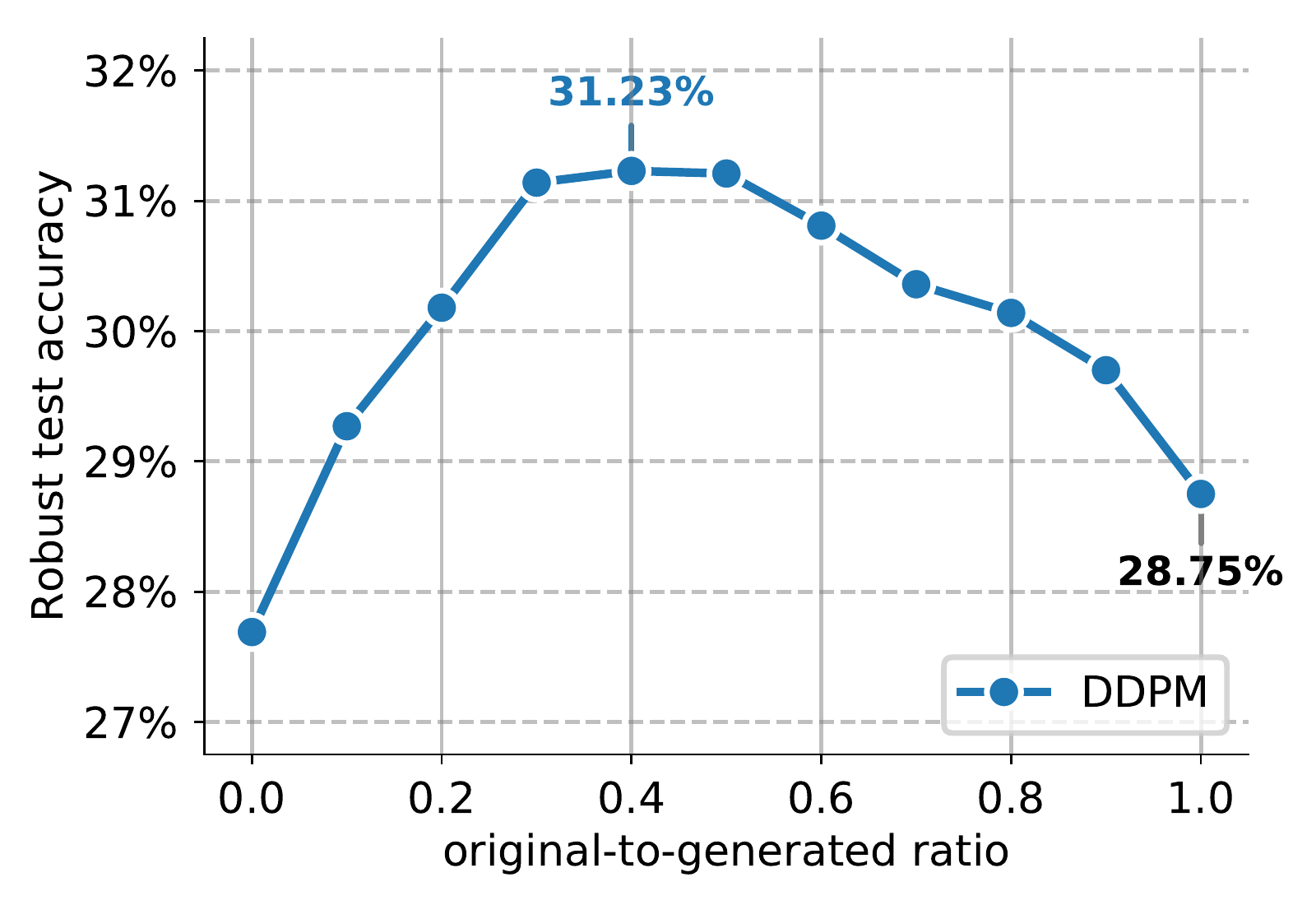}}
\subfigure[\gls{ddpm} on \svhn\label{fig:svhn_ratio}]{\includegraphics[width=.4\columnwidth]{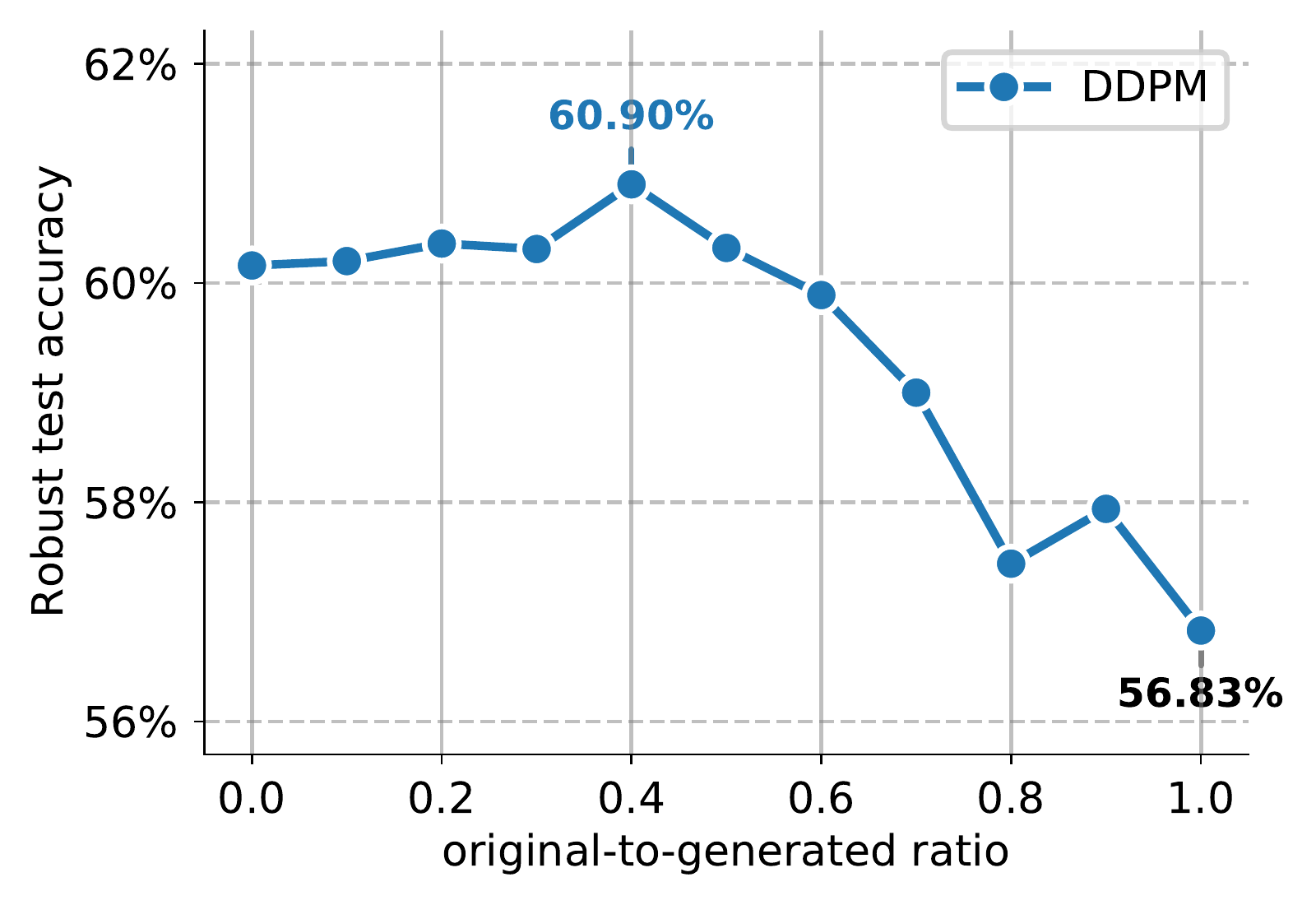}}
\caption{Robust test accuracy (against \textsc{AA+MT}) when training a \wrn-28-10 against \linf norm-bounded perturbations of size $\epsilon=8/255$ on \subref{fig:cifarh_ratio} \cifarh and \subref{fig:svhn_ratio} \svhn when using additional data produced by a \gls{ddpm}. We compare how the ratio between original images and generated images in the training minibatches affects the test robust performance (0 means generated samples only, while 1 means original \cifarh/\svhn train set only). \label{fig:more_ratio}}
\end{figure*}

For all four generative models (including the conditional Gaussian baseline), we study how the robust accuracy is impacted by the mixing ratio between original and generated images in the training minibatches.
As done in the main manuscript, \autoref{fig:generative_ratio_v2_all} explores a wide range of original-to-generated ratios (e.g., a ratio of 0.3 indicates that for every 3 original images, we include 7 generated images) while training a \wrn-28-10 against $\epsilon_\infty = 8/255$ on \cifar.
A ratio of zero indicates that only generated images are used, while a ratio of one indicates that only images from the \cifar training set are used.
We observe that samples from all models improve robustness when mixed optimally.
Surprisingly, this is the case even for the simpler conditional Gaussian baseline, which improves robust accuracy by +0.93\%.
This result provides further evidence that robustness can be improved by adding more diverse samples (even when not sampled from the original data distribution).
This is complementary to the heuristics-driven augmentations that only add data near the original training samples (as visible in \autoref{table:similarity_train_test_self}).
\autoref{fig:more_ratio} shows a similar trend for \cifarh and \svhn.

\clearpage
\section{Randomness is Enough}
\label{sec:random_is_enough}

In this section, we provide three sufficient conditions that explain why generated data helps improve robustness: \textit{(i)} the pre-trained, non-robust classifier used for pseudo-labeling (see \autoref{sec:details_data_augment}) must be accurate enough, \textit{(ii)} the likelihood of sampling examples that are adversarial to this non-robust classifier must be low, and \textit{(iii)} it is possible to sample \emph{real} images with enough frequency.

\subsection{Setup}

\def\fnr{{f_\textrm{NR}}}  %
\def\fs{{f^\star}}  %
\def\fr#1{{f^{#1}_\textrm{R}}}  %
\def\pg{{\hat{p}}}  %
\def\pt{{p^\star}}  %
\def\gXall{{\gX_\textrm{all}}}  %
\def\prob{{\mathbb{P}}}

Given access to a pre-trained \uline{non-robust} classifier $\fnr: \gXall \mapsto \gY$ and an \uline{unconditional generative} model approximating the true data distribution $\pt$ by a distribution $\pg$ over $\gXall$, we would like to train a \uline{robust} classifier $\fr{\vtheta}$ parametrized by $\vtheta$.

The set of inputs $\gXall = \{0, 1/255, \ldots, 1\}^n$ is the set of all images (discretized and scaled between 0 and 1) with dimensionality $n$.\footnote{Discretizing the image space is not necessary, but makes the mathematical notations simpler.}
The set of labels $\gY \in 2^{\mathbb{Z}}$ is a set of integers, each of which represents a given class (e.g., \emph{dog} as opposed to \emph{cat}).
There exists an image manifold $\gX \subseteq \gXall$ that contains all \emph{real} images (i.e., images for which we want to enforce robustness).
The distribution of \emph{real} images is denoted $\pt$ with $\prob_{\vx\sim\pt}(\vx) > 0$ if $\vx \in \gX$ and $\prob_{\vx\sim\pt}(\vx) = 0$ otherwise.
We further assume that each image $\vx \in \gX$ can be assigned one and only one label $y = \fs(\vx)$ where $\fs: \gX \mapsto \gY$ is a perfect classifier (only valid for \emph{real} images).
Given a perturbation ball $\sS(\vx)$\footnote{E.g., $\sS(\vx) = \{\vx' : \| \vx - \vx' \|_\infty \leq \epsilon \}$}, we restrict labels such that there exists no \emph{real} image within the perturbation ball of another that has a different label; i.e., $\forall \vx' \in \sS(\vx) \cap \gX$ we have $\fs(\vx) = \fs(\vx')$ for all $\vx \in \gX$.

Overall, we would like find optimal parameters $\vtheta^\star$ for $\fr{\vtheta^\star}$ that minimize the adversarial risk,
\begin{equation}
    \vtheta^\star = \argmin_\vtheta \E_{\vx \sim \pt} \left[ \maximize_{\vx' \in \sS(\vx)} \mathds{1}_{\fr{\vtheta}(\vx') \neq \fs(\vx)} \right]
    \label{eq:risk}
\end{equation}
without enumerating all \emph{real} images or the ideal classifier.
As such, we settle for the following sub-optimal parameters
\begin{equation}
    \hat{\vtheta}^\star = \argmin_\vtheta \E_{\vx \sim \pg} \left[ \maximize_{\vx' \in \sS(\vx)} \mathds{1}_{\fr{\vtheta}(\vx') \neq \fnr(\vx)} \right].
    \label{eq:approx_risk}
\end{equation}

\paragraph{Relationship to our method.}

The above setting corresponds to the one studied in the main manuscript where a generative model is trained on a limited number of samples from the true data distribution $\sD_\textrm{train} = \{ x_i \sim \pt \}_{i = 1}^N$.
During training, we mix \emph{real} and \emph{generated} samples and solve the following problem:
\begin{equation}
\resizebox{.94\linewidth}{!}{
    $\argmin_\vtheta \alpha \cdot \E_{\vx \in \sD} \left[ \maximize_{\vx' \in \sS(\vx)} l_\textrm{ce} \left(\fr{\vtheta}(\vx'), \fs(\vx)\right) \right] + (1 - \alpha) \cdot \E_{\vx \sim \pg'} \left[ \maximize_{\vx' \in \sS(\vx)} l_\textrm{ce} \left(\fr{\vtheta}(\vx'), f'_\textrm{NR}(\vx)\right) \right].$
    \label{eq:practical}
}
\end{equation}
where $\alpha$ is the ratio of original-to-generated samples (see \autoref{sec:generation}), $f'_\textrm{NR}$ is the underlying pre-trained classifier (used for generated samples only), $\pg'$ is the generative model distribution (which excludes samples from the train set, e.g. \gls{ddpm}) and where the 0-1 loss is replaced with the cross-entropy loss $l_\textrm{ce}$.
Ignoring the change of loss, \autoref{eq:practical} can be formulated exactly as \autoref{eq:approx_risk} by having
\begin{align}
    \fnr(x) = \begin{cases}
        \fs(\vx) & \textrm{if~} \vx \in \sD \\
        f'_\textrm{NR}(\vx) & \textrm{otherwise}
    \end{cases}
\end{align}
and by sampling a datapoint $\vx$ from the distribution of our generative model as
\begin{equation}
    \vx = \mathds{1}_{r \leq \alpha} \vx' + \mathds{1}_{r > \alpha} \vx'' \textrm{~with~} r \sim \mathcal{U}_{[0, 1]}, \vx' \sim \mathcal{U}_\sD \textrm{~and~} \vx'' \sim \pg'
\end{equation}
where $\mathcal{U}_\sA$ corresponds to the uniform distribution over set $\sA$.

\subsection{Sufficient Conditions}

In order to obtain sub-optimal parameters $\hat{\vtheta}^\star$ that approach the performance of the optimal parameters $\vtheta^\star$, the following conditions are sufficient (in the limit of infinite capacity and compute).\footnote{Understanding to which extent violations of these conditions affect robustness remains part of future work.}
These provide a deeper understanding of our method.

\paragraph{Condition 1 (accurate classifier).}
The pre-trained non-robust classifier $\fnr$ must be accurate.
When $\pt = \pg$, \autoref{eq:approx_risk} can be made identical to \autoref{eq:risk} by setting $\fnr(\vx) = \fs(\vx)$ for all $\vx \in \gX$.
For all practical settings, we posit that ``good'', sub-optimal parameters $\hat{\vtheta}^\star$ can be obtained even when the non-robust classifier $\fnr$ is not perfect.
However, it must achieve sufficient accuracy.
On \cifar, typical classifiers that are solely trained on images from the train set can reach high accuracy.
In our work, we use a pseudo-labeling classifier that achieves 95.68\% on the \cifar test set.

\paragraph{Condition 2 (unlikely attacks).}
The probability of sampling a point $\vx\sim\pg$ outside the image manifold such that it is adversarial to $\fnr$ is low:
\begin{equation}
    \prob_{\vx\sim\pg}\left(\exists \vx' \in \gX \textrm{~with~} \fnr(\vx) \neq \fnr(\vx') \textrm{~and~} \vx \in \sS(\vx')\right) < \delta, ~\delta \geq 0.
\end{equation}
To understand why this condition is needed, it is worth considering the optimal non-robust classifier $\fnr(\vx) = \fs(\vx)$ for all $\vx \in \gX$.
When $\pt \neq \pg$, it becomes possible to sample points outside the manifold of \emph{real} images and for which no correct labels exist.
Fortunately, in the limit of infinite capacity, these points can only influence the accuracy of the final robust classifier on \emph{real} images if they are within an adversarial ball $\sS(\vx)$ for a \emph{real} image $\vx$.
In practical settings, it is well documented that random sampling (e.g., using uniform or Gaussian sampling) is unlikely to produce images that are adversarial.
Hence, we posit than, unless the generative model represented by $\pg$ is trained to produce adversarial images, this condition is met.

\paragraph{Condition 3 (sufficient coverage).}
There must a non-zero probability of sampling any point on the image manifold:
\begin{equation}
   \prob_{\vx\sim\pg}(\vx) > 0, \quad \forall \vx \in \gX.
\end{equation}
In other words, the generative model should output a \emph{diverse} set of samples and some of these samples should look like \emph{real} images.
Note that it remains possible to obtain a ``good'', sub-optimal robust classifier when this condition does not hold.
However, its accuracy will rapidly decrease as coverage drops.
Hence, it is important to avoid using generative models that collapsed to a few modes and exhibit low diversity.

\subsection{Discussion}

This last condition explains why samples generated by a simple class-conditional Gaussian-fit can be used to improve robustness.
Indeed, these conditions imply that it is not necessary to have access to either the true data distribution or a perfect generative model when given enough compute and capacity.
However, it is also worth understanding what happens when capacity or compute is limited.
In this case, it is critical that the optimization focuses on \emph{real} images and that the distribution $\pg$ be as close as possible to the true distribution $\pt$.
In practice, this translates to the fact that better generative models (such as \gls{ddpm}) can be used to achieve better robustness.

\end{document}